\pdfoutput=1

\documentclass[11pt]{article}

\usepackage{ACL2023}

\usepackage{times}
\usepackage{latexsym}

\usepackage[T1]{fontenc}

\usepackage[utf8]{inputenc}

\usepackage{microtype}

\usepackage{inconsolata}
\usepackage{multirow}
\usepackage{amssymb,amsfonts,amsmath}
\usepackage{upgreek}
\usepackage{graphicx}
\usepackage{bm}

\title{Are Intermediate Layers and Labels Really Necessary? \\A General Language Model Distillation Method}

\author{
  Shicheng Tan${}^{*1}$,~
  Weng Lam Tam${}^{2}$,~
  Yuanchun Wang${}^{3}$,~
  Wenwen Gong${}^{4}$,
  \\
  {\bf Shu Zhao\footnotemark[2] ${}^{1}$},~
  {\bf Peng Zhang${}^{2}$},~
  {\bf Jie Tang\footnotemark[2] ${}^{4}$}
  \\
  ${}^{1}$Anhui University,~
  ${}^{2}$Zhipu.AI,~
  ${}^{3}$Renmin University of China,~
  ${}^{4}$Tsinghua University \\
  {\tt tsctan@foxmail.com,\{rainatam9784,frederickwang99\}@gmail.com} \\
  {\tt wenweng@mail.tsinghua.edu.cn,zhaoshuzs2002@hotmail.com} \\
  {\tt peng.zhang@zhipuai.cn,jietang@tsinghua.edu.cn} \\
}

\begin{document}
\maketitle
\renewcommand{\thefootnote}{\fnsymbol{footnote}}
\footnotetext[1]{This work was done when the author visited Zhipu.AI.}
\footnotetext[2]{Corresponding authors.}
\renewcommand{\thefootnote}{\arabic{footnote}}

\begin{abstract}
The large scale of pre-trained language models poses a challenge for their deployment on various devices, with a growing emphasis on methods to compress these models, particularly knowledge distillation.
However, current knowledge distillation methods rely on the model's intermediate layer features and the golden labels (also called hard labels), which usually require aligned model architecture and enough labeled data respectively. Moreover, the parameters of vocabulary are usually neglected in existing methods. 
To address these problems, we propose a general language model distillation (GLMD) method that performs two-stage word prediction distillation and vocabulary compression, which is simple and surprisingly shows extremely strong performance. 
Specifically, GLMD supports more general application scenarios by eliminating the constraints of dimension and structure between models and the need for labeled datasets through the absence of intermediate layers and golden labels.
Meanwhile, based on the long-tailed distribution of word frequencies in the data, GLMD designs a strategy of vocabulary compression through decreasing vocabulary size instead of dimensionality. Experimental results show that our method outperforms 25 state-of-the-art methods on the SuperGLUE benchmark, achieving an average score that surpasses the best method by 3\%.
\footnote{The code is available at \url{https://github.com/aitsc/GLMKD}.}

\end{abstract}

\section{Introduction}
The exponential increase in the scale of pre-trained language models has impeded their deployment on a wider range of devices. To mitigate the inference cost of large-scale pre-trained language models, researchers have increasingly focused on model compression methods, aiming to compress a large model into a small one with as little performance loss as possible \citep{RN7342}. While model compression can yield very small models, maintaining performance without degradation is still a challenging task, particularly when the large and small models have a significant discrepancy in parameter size \citep{RN7419}. There are various methods of model compression \citep{RN5026}, including network pruning \citep{RN7347}, quantization \citep{RN7455}, neural architecture search \citep{DBLP:journals/jmlr/ElskenMH19}, parameter sharing \citep{RN2196}, matrix decomposition \citep{RN7328}, and knowledge distillation \citep{RN7509,RN7316} etc. Currently, knowledge distillation is an important research direction, which allows for the transfer of knowledge from a large model (the teacher) to a small one (the student).

There are two main optimization objectives of the earliest knowledge distillation methods \citep{RN4609}: increasing the similarity between the student's prediction probabilities for the task and those of the teacher (soft targets); increasing the similarity between the student's predictions and the golden labels (hard targets). When the knowledge distillation method is applied to language models, there are typically two directions for improvement: leveraging the intermediate layer features of the teacher model, such as hidden states and attention, to obtain additional hidden state knowledge \citep{RN7494,RN7341}; and refining the two objectives (soft targets and hard targets) and weights of objectives \citep{RN7413,RN7311}. As shown in Table~\ref{tab:method}, these methods all rely on intermediate layer features or hard labels of the model. However, using intermediate layer features and hard labels is often accompanied by certain limitations, such as the requirement for the teacher and student models to have the same structure and dimensions, or the need for additional data and labels. These limitations make the implementation of distillation complex and hinder the applicability of these methods to a wider range of models and data. Moreover, existing methods often reduce the parameter scale of the model by decreasing the number of layers and hidden dimensions, neglecting the impact of vocabulary size.

\begin{table}
\centering
\resizebox{\linewidth}{!}{
\begin{tabular}{c|ccc} \hline
	\textbf{Methods} & \textbf{Inter} & \textbf{Soft} & \textbf{Hard} \\ \hline
    PKD \citep{RN7494} & $\checkmark$ & $\checkmark$ & $\checkmark$ \\ 
    DistilBERT \citep{RN6418} & $\checkmark$ & $\checkmark$ & $\checkmark$ \\ 
    Theseus \citep{RN7113} & $\checkmark$ &  & $\checkmark$ \\
    TinyBERT \citep{RN7482} & $\checkmark$ & $\checkmark$ &  \\
    SID \citep{RN7487} & $\checkmark$ & $\checkmark$ &  \\
    MobileBERT \citep{RN7111} & $\checkmark$ &  & $\checkmark$ \\
    CoDIR \citep{RN7473} & $\checkmark$ & $\checkmark$ & $\checkmark$ \\
    MiniLM \citep{RN7471} & $\checkmark$ &  &  \\
    MiniLMv2 \citep{RN7385} & $\checkmark$ &  &  \\
    ALP-KD \citep{RN7406} & $\checkmark$ & $\checkmark$ & $\checkmark$ \\
    LRC-BERT \citep{RN7437} & $\checkmark$ & $\checkmark$ & $\checkmark$ \\
    Annealing-KD \citep{RN5665} &  & $\checkmark$ & $\checkmark$ \\
    CKD \citep{RN7407} & $\checkmark$ & $\checkmark$ & $\checkmark$ \\
    Universal-KD \citep{RN7376} & $\checkmark$ & $\checkmark$ & $\checkmark$ \\
    Meta-KD \citep{RN7410} & $\checkmark$ & $\checkmark$ &  \\
    HRKD \citep{RN7445} & $\checkmark$ & $\checkmark$ &  \\
    RW-KD \citep{RN7413} &  & $\checkmark$ & $\checkmark$ \\
    MetaDistil \citep{RN7311} &  & $\checkmark$ & $\checkmark$ \\
    DIITO \citep{RN7325} & $\checkmark$ & $\checkmark$ & $\checkmark$ \\
    Continuation-KD \citep{Continuation} &  & $\checkmark$ & $\checkmark$ \\
    RAIL-KD \citep{RN7350} & $\checkmark$ & $\checkmark$ & $\checkmark$ \\
    MGSKD \citep{RN7341} & $\checkmark$ & $\checkmark$ &  \\
    TMKD \citep{RN6109} &  & $\checkmark$ & $\checkmark$ \\
    MT-BERT \citep{RN7379} & $\checkmark$ & $\checkmark$ & $\checkmark$ \\
    RL-KD \citep{RN7370} &  & $\checkmark$ & $\checkmark$ \\
    Uncertainty \citep{RN7419} &  & $\checkmark$ &  $\checkmark$\\
    \hline
\end{tabular}
}
\caption{Almost all knowledge distillation methods for language models are based on either intermediate layer features (Inter) or hard labels (Hard). Soft denotes soft labels, specifically, the logits of the teacher model in downstream task loss.}
\label{tab:method}
\end{table}

To address these problems, we propose a general language model distillation (GLMD) method that performs two-stage (pre-training and task-specific stages) word prediction distillation and vocabulary compression.
Specifically, GLMD distills the model using only the language modeling word prediction logits during the pre-training stage, which is similar to the soft labels used in general methods. The key to this stage is that we distill both masked and unmasked tokens. In the task-specific stage (fine-tuning), GLMD distills both the language modeling word prediction logits and the soft labels. The language modeling word prediction logits is crucial in this stage, making the distillation more consistent between the pre-training and task-specific stages. In these two stages, GLMD eliminates the need for complicated intermediate layers and golden labels and does not require the selection of intermediate layers or labeled dataset. Meanwhile, GLMD uses the teacher vocabulary to map low-frequency words to the most similar high-frequency words, further compressing the model with  almost no performance loss.

In summary, our major contributions are:

\begin{itemize}
\item We propose a general language model distillation (GLMD) method that saves the tedious work on intermediate layer features and golden labels, and does not require the selection of intermediate layers or labeled dataset. We demonstrate through analysis that GLMD allows models to autonomously learn intermediate layer features that are similar to those of the teacher.
\item We propose a vocabulary compression strategy based on the long-tailed distribution of words in data, which reduces the vocabulary size without reducing dimensions of the model. Additionally, our vocabulary compression strategy can be used in conjunction with other dimensionality reduction strategies with very little performance loss.
\item We verify that GLMD outperforms 25 state-of-the-art model distillation methods on the SuperGLUE benchmark, achieving an average score that surpasses the best method by 3\%. Furthermore, our vocabulary compression strategy also outperforms other 2 dimensionality reduction strategies. We also investigate distillation of ultra-large-scale language models (10B-scale) for the first time.
\end{itemize}

\section{Related work}
\paragraph{Language Model Distillation}
Since the introduction of knowledge distillation to pre-trained language models by PKD \citep{RN7494}, an increasing number of researchers have recognized the importance of knowledge distillation. 
During the early stage of the research, PD \citep{RN6581} employed simple baseline (soft targets) distillation for language models, resulting in a relatively limited transfer of knowledge for the model. 
Subsequent research had primarily focused on the use of intermediate layer features in language models \citep{RN7113,RN6418}, including distillation of models during pre-training stage \citep{RN7111}, task-specific stage \citep{RN7487}, and two-stage \citep{RN7482} approaches. 
Given the typically large amount of intermediate layer features, some work had utilized features from only a single intermediate layer \citep{RN7471,RN7385}, while other work had examined methods for reducing the scale of features \citep{RN7414}.
Recent work has explored ways to utilize better intermediate layer features, for example, CoDIR \citep{RN7473} and LRC-BERT \citep{RN7437} utilized cross-sample feature relationships through contrastive learning; ALP-KD \citep{RN7406} and Universal-KD \citep{RN7376} combined all intermediate layer features through attention mechanisms; Meta-KD \citep{RN7410} and HRKD \citep{RN7445} used meta-learning to assign appropriate weights to intermediate layer features; RAIL-KD \citep{RN7350} randomly selected different intermediate layers for distillation; CKD \citep{RN7407} and MGSKD \citep{RN7341} used some variety of similarity calculation methods for intermediate layer features; DIITO \citep{RN7325} allowed student models to learn counterfactual outputs by swapping intermediate layer features between different samples. 

However, the use of intermediate layer features has additional limitations, such as requiring the same model structure \citep{RN7111} for both teacher and student, or requiring linear transformations \citep{RN7482} to ensure consistency in dimensions between teacher and student.
There were also methods that only used soft and hard targets, for example, Annealing-KD \citep{RN5665} and Continuation-KD \citep{Continuation} gradually increased the weight of soft targets through simulated annealing; RW-KD \citep{RN7413} adjusted the weight of soft and hard targets through meta-learning and a dev set; MetaDistil \citep{RN7311} allowed the teacher to learn how to output better soft labels through meta-learning and a quiz set. These approaches relied on hard labels and may have even required additional datasets for partitioning.
Additionally, there had been approaches that distilled multiple teachers \citep{RN6109,RN7379,RN7370,RN7419} or teacher assistants \citep{RN7475,RN7395} at the same time, but they still relied on intermediate layer features or hard labels. In comparison, GLMD can achieve the strongest performance in a more broadly applicable context without intermediate layer features or hard labels.

\paragraph{Vocabulary Compression}
Vocabulary compression refers to reducing the parameter size of the vocabulary in a language model.
In the knowledge distillation of language models, reducing the parameter size of the model is mainly used to reduce the number of model layers or dimensions \citep{RN7482,RN7385}.
MobileBERT \citep{RN7111} and ALBERT \citep{RN2196} independently reduced the dimensions of the vocabulary to achieve vocabulary compression. MobileBERT needed to restore the dimension of the vocabulary in the calculation of the pre-training loss due to the requirement to ensure consistency between the vocabulary dimension and the model output dimension. On the other hand, ALBERT used a linear layer to alter the output dimension of the model.
However, these vocabulary compression methods only reduced the dimensionality and ultimately required dimensionality restoration. In contrast, our vocabulary compression method reduces the number of words through mapping, further compressing the model with almost no impact on performance.

\section{Preliminaries}
In this section, we introduce the objective function for knowledge distillation of language models, and formalize the language modeling word prediction logits.

\subsection{Knowledge Distillation}
Knowledge distillation aims to transfer the knowledge of the teacher $T$ to the student $S$. The knowledge and transfer method can be formalized as model features and distance metrics respectively. Formally, knowledge distillation for language models typically consists of the following three objective functions:
\begin{equation}
\begin{array}{l}
\mathcal{L}_{\mathrm{soft}}=\uptau^2KL(\sigma({f_l^S(\mathbf x)}/{\uptau}), \sigma({f_l^T(\mathbf x)}/{\uptau})) \vspace{1ex} \\
\mathcal{L}_{\mathrm{hard}}=CE(\sigma({f_l^S(\mathbf x)}), \mathbf y) \vspace{1ex} \\
\mathcal{L}_{\mathrm{inter}}=d(f^S(\mathbf H^S), f^T(\mathbf H^T))
\end{array} 
\end{equation}
where $\uptau$ denotes the softening parameter (temperature), $KL(\cdot,\cdot)$ denotes the KL divergence, $\sigma$ denotes the softmax function, $\mathbf x\in\mathbb{R}^{l}$ denotes the input sequence (token ids) of length $l$ for the language model, $f_l^S(\mathbf x)$ and $f_l^T(\mathbf x)$ denote the logits output by the student and the teacher before computing the task loss respectively, $CE(\cdot,\cdot)$ denotes the cross entropy, $\mathbf y$ denotes the hard labels, $d(\cdot)$ denotes the distance metric (e.g., KL divergence and mean square error), $\mathbf H^S$ and $\mathbf H^T$ denote the intermediate layer features (e.g., hidden states and attention) of the student and the teacher respectively, $f^S(\cdot)$ and $f^T(\cdot)$ denote custom transformations (e.g. linear transformations) of the student and teacher features, respectively.

Currently, mainstream methods employ different combinations and weighting schemes of the three objective functions in the pre-training and task-specific stages. For example, TinyBERT \citep{RN7482} optimizes $\mathcal{L}_{\mathrm{inter}}$ in the pre-training stage and optimizes $\mathcal{L}_{\mathrm{inter}}$ and $\mathcal{L}_{\mathrm{soft}}$ in the task-specific stage, while MetaDistil \citep{RN7311} only optimizes $\mathcal{L}_{\mathrm{soft}}$ and $\mathcal{L}_{\mathrm{hard}}$ in the task-specific stage. Notably, to ensure feature dimension matching between the teacher and student, $\mathcal{L}_{\mathrm{inter}}$ relies on complex custom transformations, such as linear transformations ($f(\mathbf H)=\mathbf W\mathbf H$) and pair-wise scaled dot-product ($f(\mathbf H)=\mathbf H\mathbf H^T/\sqrt{dimensionality}$). In contrast, our method does not rely on $\mathcal{L}_{\mathrm{inter}}$ and $\mathcal{L}_{\mathrm{hard}}$.

\subsection{Language Modeling Word Prediction Logits}
Language modeling typically refers to unsupervised tasks in the pre-training stage, such as causal language modeling for GPT \citep{RN1153}, masked language modeling for BERT \citep{RN289}, and autoregressive blank filling for GLM \citep{RN2172}. This process typically requires a decoder to decode the model's output into a prediction logits for each word. The decoder is typically a linear transformation using the vocabulary parameters as weights. The language modeling word prediction logits can be formulated as follows:
\begin{equation}
\label{eq:decoder}
LM(\mathbf x)=f_t(\mathbf x)\mathbf{W}_v^T
\end{equation}
where $\mathbf{W}_v\in\mathbb{R}^{v\times h}$ denotes the vocabulary parameters (weight of embeddings-layer), and $f_t(\mathbf x)\in\mathbb{R}^{l\times h}$ denotes the output of the final layer of transformer. The scalar value $l$ denotes the length of the text sequence, $v$ denotes the number of tokens in the vocabulary, and $h$ denotes the dimensionality of the hidden layer. It is worth noting that the $LM(\cdot)$ can also be computed at the task-specific stage.

\section{Method}
\begin{figure}
  \centering
  \includegraphics[width=\linewidth,scale=1]{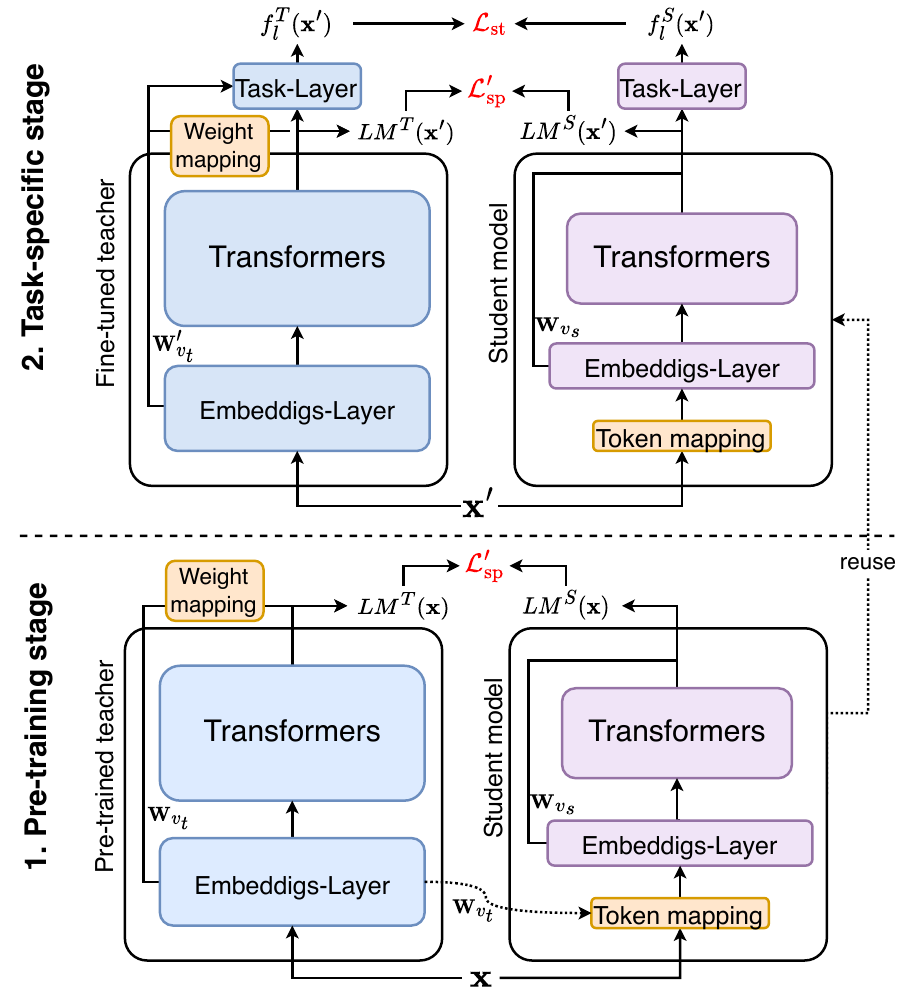}
  \caption{The framework of the GLMD. The task-layer aims to convert the $f_t(\cdot)$ and $\mathbf{W}_v$ into the logits $f_l(\cdot)$ of downstream task loss. \label{fig:two-stage}}
\end{figure}

In this section, we propose a general language model distillation (GLMD) method with two-stage word prediction distillation and vocabulary compression. Figure~\ref{fig:two-stage} shows the overview framework of GLMD, which implements a vocabulary compression strategy while performing a two-stage word prediction distillation process. We next provide a detailed description of these two components.

\subsection{Two-stage Word Prediction Distillation}
To eliminate the reliance on intermediate layer features and hard labels in distillation, we propose a two-stage word prediction distillation process based on the language modeling word prediction logits. It allows teacher models and students models to have different model structures and does not need the selection of intermediate layers. This process makes the distillation goal more closely aligned with the model's task and makes the model's distillation more consistent across both the pre-training and task-specific stages. During the pre-training stage, we optimize the student model with the objective function $\mathcal{L}^{\prime}_{\mathrm{sp}}$. We then optimize the student again using $\mathcal{L}^{\prime}_{\mathrm{sp}}$ during the task-specific stage. After these two training phases, we finally optimize the student with the objective function $\mathcal{L}_{\mathrm{st}}$ from the task-specific stage. 
Our objective functions $\mathcal{L}_{\mathrm{st}}$ and $\mathcal{L}^{\prime}_{\mathrm{sp}}$ are defined as:
\begin{equation}
\begin{array}{l}
\mathcal{L}_{\mathrm{st}}=\mathcal{L}_{\mathrm{soft}} \vspace{1ex} \\
\mathcal{L}^{\prime}_{\mathrm{sp}}=\mathcal{L}_{\mathrm{sp}}\odot\mathbf m_p \vspace{1ex} \\
\mathcal{L}_{\mathrm{sp}}=\uptau^2KL(\sigma(\frac{LM^S(\mathbf x)}{\uptau}), \sigma(\frac{LM^T(\mathbf x)}{\uptau}))
\end{array} 
\end{equation}
where $\odot$ denotes the Hadamard product, 
and $\mathbf m_p\in\mathbb{R}^{l}$ denotes the mask vector, which only masks the pad while preserving the masked and unmasked tokens. We find that unmasked tokens are typically not predicted by language modeling, but they can provide more knowledge for distillation.

$\mathcal{L}^{\prime}_{\mathrm{sp}}$ and $\mathcal{L}_{\mathrm{st}}$ represent the soft targets for the pre-training and task-specific stages, respectively. It is worth noting that $\mathcal{L}^{\prime}_{\mathrm{sp}}$ can be used in both pre-training and task-specific stages, making the optimization objectives more consistent across the two stages.

\subsection{Vocabulary Compression}
To further compress the parameter scale of the model, we propose a vocabulary compression strategy that reduces the number of tokens in the vocabulary. Because word frequencies have a long-tailed distribution, some low-frequency words can still be understood by the language model after being replaced with similar words. Let the compression rate of the vocabulary be $r_v$, and the number of all tokens before compression be $v$. We sort the tokens in the pre-trained corpus according to their frequency of occurrence. The tokens ranking in the top $vr_v$ are treated as the compressed tokens $\mathbf{w}_c\in\mathbb{R}^{vr_v}$, while the remaining tokens $\mathbf{w}_m\in\mathbb{R}^{v(1-r_v)}$ are to be mapped. Figure~\ref{fig:two-stage} illustrates the key aspect of our vocabulary compression strategy, which includes replacing low-frequency words with similar words through \textbf{token mapping} and aligning the weight matrix of the embedding-layer through \textbf{weight mapping}. The token mapping aims to map $w_{m}\in\mathbf{w}_m$ to $w_c\in\mathbf{w}_c$, and the mapping function is defined as:
\begin{equation}
f_{tm}(w_{m})=\mathop{\operatorname{argmax}}_{w_c\in\mathbf{w}_c} (Sim(v(w_m),v(w_c)))
\end{equation}
where $v(w_m)$ and $v(w_c)$ denote the token vectors of $w_m$ and $w_c$ in the pre-trained teacher's vocabulary weight $\mathbf{W}_{v_t}$, respectively. $Sim(\cdot,\cdot)$ is a function of calculating similarity using the inner product, which is similar to the decoding in Equation~\ref{eq:decoder}.
The weight mapping aims to remove $\mathbf w_m$ from $\mathbf W_{v_t}$, and the mapping function is defined as:
\begin{equation}
f_{wm}(\mathbf W_{v_t})=\mathbf W_{v_t}[\mathbf w_c]
\end{equation}
where $[\cdot]$ denotes a slicing operation, specifically obtaining all vector of $w_c\in\mathbf{w}_c$ from $\mathbf W_{v_t}$.

\section{Experiments}
In this section, we demonstrate the effectiveness of the GLMD on models with different parameter scales (110M, 340M, and 10B) and analyze the role of different components and why they are effective. All experiments were conducted on 40 NVIDIA A100 GPUs and completed within 4 months, utilizing the PyTorch framework.

\subsection{Experimental Setup}
\paragraph{Datasets}
To evaluate our GLMD method, we conduct experiments on the more challenging SuperGLUE \citep{RN2189} benchmark instead of GLUE \citep{RN2190}. The use of more difficult tasks allows for a better display of the discrepancy between different distillation methods. We use the average score across 8 tasks in SuperGLUE as the evaluation metric. We use BooksCorpus \citep{books} and English Wikipedia as the data (19GB) for the distillation of pre-training stage for all methods.

\paragraph{Baselines}
We compare 25 commonly used distillation methods as listed in Table~\ref{tab:results}. 
We provide a more detailed description of these methods in Appendix~\ref{sec:appendix:baselines}.

\begin{table}
\centering
\resizebox{\linewidth}{!}{
\begin{tabular}{l|ccc} \hline
	\textbf{Models} & \textbf{\#Params} & \textbf{\#Dimensions} & \textbf{\#Layers} \\ \hline
    $T_1$ & 110M & 768 & 12 \\ 
    $T_1$ for MobileBERT & 293M & 1024 & 24 \\ 
    $T_2$ & 340M & 1024 & 24 \\ 
    $T_3$ & 10B & 4096 & 48 \\ 
    $A_1$ & 200M & 896 & 14 \\ 
    $A_2$ & 110M & 768 & 12 \\ 
    $S_1$ & 22M & 384 & 6 \\ 
    $S_1$ for MobileBERT & 25M & 128 & 24 \\ 
    $S_2$ & 66M & 768 & 6 \\ 
    $S_3$ & 2B & 2048 & 36 \\ 
    \hline
\end{tabular}
}
\caption{The parameter sizes, hidden layer dimensions, and number of transformer layers of the teacher models ($T_1,T_2,T_3$), teacher assistant models ($A_1,A_2$), and student models ($S_1,S_2,S_3$) used by all methods.}
\label{tab:models}
\end{table}
\paragraph{Language Models}
Student and teacher models of all methods have the standard GLM \citep{RN2172} architecture. GLM (General Language Model) is a more advanced language model that inherits the advantages of both autoencoding and autoregression. We choose GLM for two reasons: it performs stronger than the commonly used BERT \citep{RN289} and RoBERTa \citep{RN1181}; it has open-source pre-trained language models with 10B-scale or even 100B-scale parameters \citep{glm130b}. All pre-trained models, with the exception of MobileBERT, which was trained by us based on GLM, were obtained from the official GLM website \footnote{https://github.com/THUDM/GLM}. Both the teacher and student models were trained with half-precision floating-point (fp16). The model sizes used in this paper are shown in Table~\ref{tab:models}.

\paragraph{Hyperparameters}
For our method, the temperatures $\uptau$ for the loss $\mathcal{L}^{\prime}_{\mathrm{sp}}$ and $\mathcal{L}_{\mathrm{st}}$ are set to 15 and 1, respectively. All baselines use the best parameters from their respective papers. For all methods that rely on the pre-training stage, the batch size, peak learning rate, and number of iterations are set to 64, 4e-4, and 150000, respectively. For all single-teacher methods, we use grid search to find the best parameters during the task-specific stage, including the learning rate \{5e-6,1e-5,2e-5\} and batch size \{16,32\}. The multi-teacher and teacher assistant methods are similar to the single-teacher methods in the core method, with differences in the weighting and assistant of teachers. The other parameters for the task-specific stage are kept consistent with the fine-tuned teacher, using the best parameters provided by GLM. The results for all experiments (w/o $T_3$-$S_3$) are the average of 3 random seeds. For more details on the hyperparameters, refer to Appendix~\ref{sec:appendix:hyper}.

\begin{table}
\centering
\resizebox{\linewidth}{!}{
\begin{tabular}{l|ccc} \hline
	\textbf{Methods} & \multicolumn{3}{c}{\textbf{SuperGLUE}}\\ \hline
    \textbf{\textit{Fine-tuned teacher}} & $T_1$ & $T_2$ & $T_3$ \\
    GLM \citep{RN2172} & 71.7 & 77.1 & 89.0 \\ \hline
    \textbf{\textit{Single-teacher}} & $T_1$-$S_1$ & $T_1$-$S_2$ & $T_2$-$S_2$ \\
    \quad (compression ratio) & (0.2) & (0.5) & (0.2) \\
    \quad (parameter size of the student) & (22M) & (66M) & (66M) \\
    KD \citep{RN4609} & 52.0 & 52.4 & 53.2 \\ 
    PD \citep{RN6581} & 61.0 & 62.0 & 61.4 \\ 
    PKD \citep{RN7494} & - & 66.2 & - \\ 
    DistilBERT \citep{RN6418} & - & 63.8 & - \\ 
    Theseus \citep{RN7113} & - & 66.1 & - \\
    TinyBERT \citep{RN7482} & 67.3 & 70.8 & 69.1 \\
    SID \citep{RN7487} & - & 55.1 & - \\
    MobileBERT$_\mathrm{25M}$ \citep{RN7111} & 65.1 & - & - \\
    MiniLM \citep{RN7471} & 61.6 & 63.8 & 62.4 \\
    MiniLMv2 \citep{RN7385} & 62.1 & 63.7 & 66.1 \\
    ALP-KD \citep{RN7406} & - & 64.3 & - \\
    LRC-BERT \citep{RN7437} & 58.4 & 62.3 & 60.8 \\
    Annealing-KD \citep{RN5665} & 61.0 & 63.3 & 63.1 \\
    CKD \citep{RN7407} & 61.9 & 63.3 & 62.6 \\
    Universal-KD \citep{RN7376} & 52.9 & 66.2 & 54.5 \\
    DIITO \citep{RN7325} & - & 66.8 & - \\
    Continuation-KD \citep{Continuation} & 61.0 & 62.7 & 61.7 \\
    RAIL-KD \citep{RN7350} & 61.2 & 63.6 & 60.4 \\
    MGSKD \citep{RN7341} & 58.8 & 63.5 & 59.9 \\
    GLMD$_{\mathrm{-vc}}$ \textit{(ours)} & \bf 67.9 & \bf 71.5 & 72.1 \\
    GLMD \textit{(ours)} & 67.4 & 70.9 & \bf 72.2 \\ 
    \quad (parameter size of the student) & (16M) & (55M) & (55M) \\ \hline
    \textbf{\textit{Single-teacher (10B-scale)}} & \multicolumn{3}{c}{$T_3$-$S_3$ (2B)} \\ 
    TinyBERT \citep{RN7482} & \multicolumn{3}{c}{65.09 (w/o ReCoRD)} \\
    GLMD$_{\mathrm{-vc}}$ \textit{(ours)} & \multicolumn{3}{c}{\textbf{84.76} (w/o ReCoRD)} \\ \hline
    \textbf{\textit{Multi-teacher}} & \multicolumn{3}{c}{$(T_1,T_2)$-$S_2$ (66M)} \\ 
    TMKD \citep{RN6109} & \multicolumn{3}{c}{65.6} \\
    MT-BERT \citep{RN7379} & \multicolumn{3}{c}{59.1} \\
    RL-KD \citep{RN7370} & \multicolumn{3}{c}{65.3} \\
    Uncertainty \citep{RN7419} & \multicolumn{3}{c}{65.4} \\ \hline
    \textbf{\textit{Teacher assistants}} & \multicolumn{3}{c}{$T_2$-$A_1$-$A_2$-$S_2$ (66M)} \\ 
    TAKD \citep{RN7475} & \multicolumn{3}{c}{54.5} \\
    DGKD \citep{RN7395} & \multicolumn{3}{c}{54.0} \\ \hline
    \textbf{\textit{Vocabulary compression}} & \multicolumn{3}{c}{$T_1$-$S_1$,~$r_v=0.5$ (16M)} \\ 
    GLMD$_{\mathrm{-vc+mo}}$ \citep{RN7111} & \multicolumn{3}{c}{67.0} \\
    GLMD$_{\mathrm{-vc+al}}$ \citep{RN2196} & \multicolumn{3}{c}{67.3} \\
    GLMD$_{\mathrm{+al}}$ \textit{(ours)} (14M) & \multicolumn{3}{c}{\bf 67.4} \\
    \hline
\end{tabular}
}
\caption{The average scores of all methods on the SuperGLUE dev set. 
The inference speed is essentially proportional to the scale of the student's parameters. 
Detailed results for each dataset of SuperGLUE can be found in Appendix~\ref{sec:appendix:results}.}
\label{tab:results}
\end{table}

\begin{table}
\centering
\resizebox{\linewidth}{!}{
\begin{tabular}{l|cc} \hline
	\textbf{Methods} & \textbf{SuperGLUE} \\ \hline
    \textbf{\textit{Two-stage word prediction distillation}} & $T_1$-$S_1$ \\ 
    GLMD$_{\mathrm{-vc}}$ & \bf 67.9 \\ 
    \quad w/o $\mathbf m_p$ in $\mathcal{L}^{\prime}_{\mathrm{sp}}$ & 67.3 \\ 
    \quad w/o unmasked tokens in $\mathbf m_p$ of $\mathcal{L}^{\prime}_{\mathrm{sp}}$ & 66.2 \\ 
    \quad w/o $\mathcal{L}^{\prime}_{\mathrm{sp}}$ in task-specific stage & 64.4 \\
    \quad add same inter loss as TinyBERT & 67.2 \\
    \quad add $\mathcal{L}_{hard}$ in task-specific stage & 67.5 \\
    \quad replace $KL$ with $MSE$ in $\mathcal{L}^{\prime}_{\mathrm{sp}}$ & 66.7 \\
    \quad replace $KL$ with $MSE$ in $\mathcal{L}_{\mathrm{st}}$ & 66.7 \\ \hline
    \textbf{\textit{Vocabulary compression}} & $T_1$-$S_1$,$r_v=0.5$ \\ 
    GLMD & \bf 67.4 \\ 
    \quad replace $Sim(\cdot)$ with Cosine similarity & 66.8 \\ 
    \quad replace $Sim(\cdot)$ with - Euclidean distance & 66.3 \\ 
    \quad replace $f_{map}(\cdot)$ with [UNK] token id & 65.1 \\ 
    \quad add token mapping for teacher & 65.5 \\
    \hline
\end{tabular}
}
\caption{Ablation study on the SuperGLUE dev set.}
\label{tab:ablation}
\end{table}

\subsection{Main Results}
In Table~\ref{tab:results}, we report the average scores of all methods on the SuperGLUE dev set. GLMD$_{\mathrm{-vc}}$ denotes GLMD without vocabulary compression strategy. GLMD$_{\mathrm{-vc+mo}}$ and GLMD$_{\mathrm{-vc+al}}$ denote the use of MobileBERT and ALBERT vocabulary compression strategies on GLMD, respectively. GLMD$_{\mathrm{+al}}$ denotes the combination of ALBERT and our vocabulary compression strategies on GLMD.

GLMD achieves the highest performance among 25 baselines on $T_1$-$S_1$, $T_1$-$S_2$, and $T_2$-$S_2$ scales, with a 0.1\%, 0.1\%, and 3.1\% improvement over the best method (TinyBERT), respectively. More importantly, in a fair environment without vocabulary compression, GLMD$_{\mathrm{-vc}}$ outperforms the best method by 0.7\%, 0.7\%, and 3.0\%, respectively. This demonstrates that high-performance distillation does not necessarily require intermediate layer features or hard labels, whether reducing the number of layers or the dimensionality of the student model.
GLMD significantly outperforms TinyBERT in the distillation process on the scale of 10B to 2B, indicating that TinyBERT is not suitable for ultra-large-scale model distillation on the SuperGLUE benchmark.
The use of vocabulary compression in GLMD still maintains strong competitiveness in further compressing the model. GLMD outperforms the best vocabulary compression strategy (GLMD$_{\mathrm{-vc+al}}$) by 0.1\% on $T_1$-$S_1$ scale, confirming that reducing the vocabulary size is an effective strategy. It is worth noting that our vocabulary compression strategy can be combined with other dimensionality reduction methods, such as GLMD$_{\mathrm{+al}}$, which can maintain the original performance even with only one-fourth of the vocabulary parameters. Additionally, some recent baselines did not show the strongest performance, we discuss more factors affecting baseline performance in appendix~\ref{sec:appendix:analysis}.

\subsection{Ablation Study}
After having validated the effectiveness of GLMD and GLMD$_{\mathrm{-vc}}$, we further analyze in Table~\ref{tab:ablation} the key design factors that impact the performance of the two components in greater detail.
\textbf{(1) Two-stage word prediction distillation.} The results indicate that both removing $\mathbf m_p$ (row 4) or removing unmasked tokens (row 5) from $\mathbf m_p$ do not perform as well as GLMD$_{\mathrm{-vc}}$ (row 3), which confirms the effectiveness of $\mathbf m_p$ in $\mathcal{L}^{\prime}_{\mathrm{sp}}$. The use of $\mathbf m_p$ in $\mathcal{L}^{\prime}_{\mathrm{sp}}$ in the task-specific stage makes the distillation of the student more consistent in the pre-training and task-specific stages, which is verified by row 6. The performance degradation observed upon incorporating intermediate layer features (row 7) or hard labels (row 8) into the loss function in GLMD$_{\mathrm{-vc}}$ further confirms that such features and labels are not necessary. Additionally, we find that the KL divergence performed better than the MSE  (mean square error) in both $\mathcal{L}^{\prime}_{\mathrm{sp}}$ and $\mathcal{L}_{\mathrm{st}}$ (rows 9 and 10).
\textbf{(2) Vocabulary compression.} In addition to mapping low-frequency tokens to similar tokens using the decoder approach, we also attempt to use Cosine similarity (row 13), Euclidean distance (row 14), and direct replacement with [UNK] (row 15) to map similar tokens. We found that these mapping methods did not perform as well as GLMD (row 12), which may be because the mapping method used in GLMD is closer to the decoding approach used in language modeling task. The result of line 12 outperforming line 16 verifies that token mapping is only applicable for students.

\section{Analysis}

\begin{figure}[!t]
  \centering
  \includegraphics[width=\linewidth,scale=1]{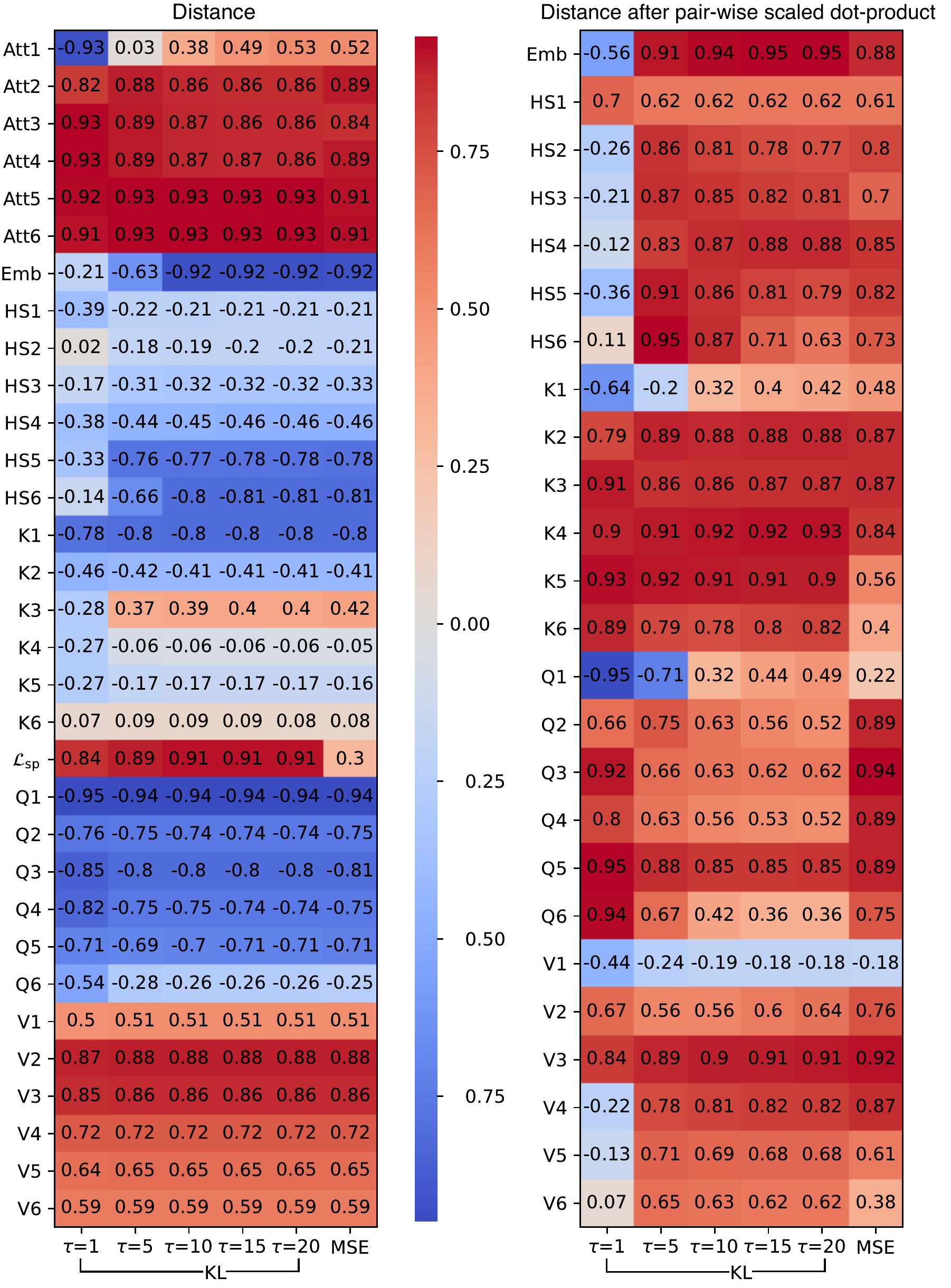}
  \caption{Spearman correlation coefficient of the $\mathcal{L}^{\prime}_{\mathrm{sp}}$ ($\uptau=15$) with the distance between teacher and student features during pre-training stage of GLMD$_{\mathrm{-vc}}$ ($T_1$-$S_2$). Let the intermediate feature be $\mathbf H\in\mathbb{R}^{l\times h}$, and the distance after pair-wise scaled dot-product is calculated by first computing $f(\mathbf H)=\frac{\mathbf H\mathbf H^T}{\sqrt{h}}$. HS1, Att1, Q1, K1, and V1 denote the hidden state, attention scores, query, key, and value of the first layer transformer, respectively. Emb denotes the output of the embedding-layer. \label{fig:glmd-analysis}}
\end{figure}

\begin{figure*}
  \centering
  \includegraphics[width=\linewidth]{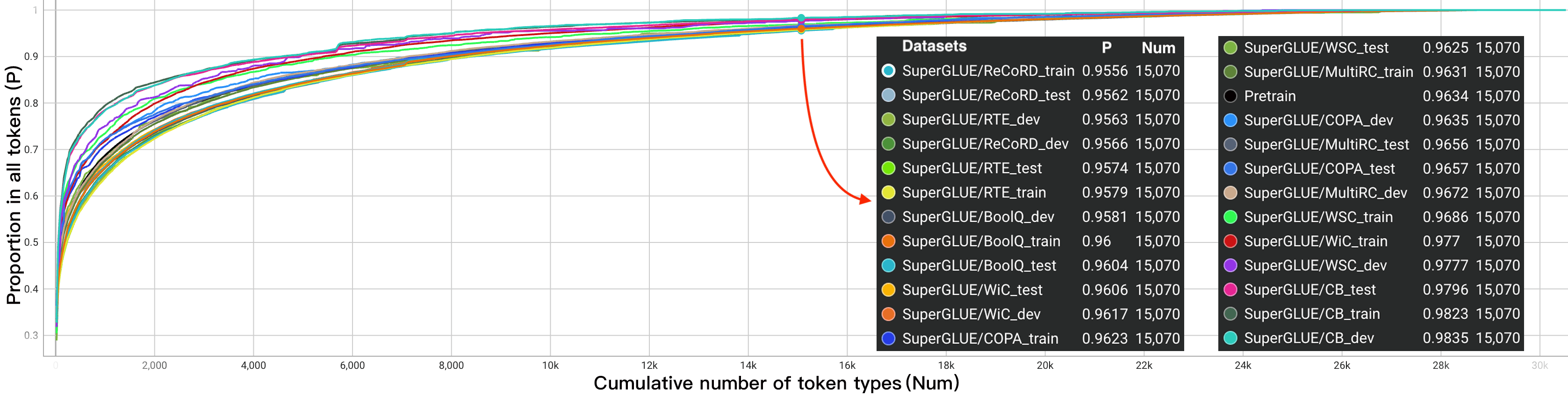}
  \caption{At least half of the tokens can cover 95.56\% of the dataset corpus. We count the number of times a token appeared in the pre-trained corpus and sorted them in decreasing order on the x-axis. The y-axis shows the proportion of the 8 datasets (divided into train/dev/test) in SuperGLUE benchmark that can be covered by x tokens. \label{fig:tokens}}
\end{figure*}

In this section, we analyze the reasons behind the work of GLMD and the impact of hyperparameters on performance in GLMD.
\subsection{Why Does GLMD Work?}
\label{analysis:why}
Compared to methods using only soft or hard labels, the $\mathcal{L}^{\prime}_{\mathrm{sp}}$ in GLMD$_{\mathrm{-vc}}$ clearly provides more knowledge, but it is still unclear why the intermediate feature is not necessary. We hypothesize that $\mathcal{L}^{\prime}_{\mathrm{sp}}$ reduces inductive bias and allows the model to spontaneously learn intermediate features that should be similar to the teacher. To verify this hypothesis, we calculate the spearman correlation between the distance $d(f^S(\mathbf H^S), f^T(\mathbf H^T))$ and $\mathcal{L}^{\prime}_{\mathrm{sp}}$ during pre-training stage of GLMD$_{\mathrm{-vc}}$. The red part in Figure~\ref{fig:glmd-analysis} shows that as $\mathcal{L}^{\prime}_{\mathrm{sp}}$ decreases, not all the distance of features between teacher and student is getting close during distillation so that it may not be necessary to draw all the intermediate features close as in existed methods, supporting our hypothesis.

We hypothesize that the success of the vocabulary compression strategy is based on the long-tail distribution of tokens, where some low-frequency tokens can still be understood by the language model after being replaced with similar tokens. Figure~\ref{fig:tokens} verifies the long-tail distribution of tokens. The result in row 16 of Table~\ref{tab:ablation} shows that using token mapping for teacher results in a decrease in performance. This verifies that even when some low-frequency tokens are replaced with similar tokens, students can still learn the meaning of these tokens from teachers without token mapping.

\begin{table}[h!t]
\centering
\resizebox{\linewidth}{!}{
\begin{tabular}{l|cc} \hline
	\textbf{Methods} & \textbf{SuperGLUE} \\ \hline
    \textbf{\textit{Two-stage word prediction distillation}} & $T_1$-$S_1$ \\ 
    GLMD$_{\mathrm{-vc}}$ & \bf 67.9 \\ 
    \quad set $\uptau=1$ in $\mathcal{L}^{\prime}_{\mathrm{sp}}$ & 65.0 \\ 
    \quad set $\uptau=5$ in $\mathcal{L}^{\prime}_{\mathrm{sp}}$ & 67.0 \\ 
    \quad set $\uptau=10$ in $\mathcal{L}^{\prime}_{\mathrm{sp}}$ & 66.9 \\ 
    \quad set $\uptau=20$ in $\mathcal{L}^{\prime}_{\mathrm{sp}}$ & 67.6 \\ 
    \quad set $\uptau=5$ in $\mathcal{L}_{\mathrm{st}}$ & 67.5 \\ 
    \quad set $\uptau=10$ in $\mathcal{L}_{\mathrm{st}}$ & 67.5 \\ 
    \quad set $\uptau=15$ in $\mathcal{L}_{\mathrm{st}}$ & 67.5 \\ 
    \quad set $\uptau=20$ in $\mathcal{L}_{\mathrm{st}}$ & 67.5 \\ 
    \quad set $\uptau=100$ in $\mathcal{L}_{\mathrm{st}}$ & 67.8 \\ 
    \quad set $\uptau=200$ in $\mathcal{L}_{\mathrm{st}}$ & 67.4 \\ 
    \quad set $\uptau=1000$ in $\mathcal{L}_{\mathrm{st}}$ & 67.6 \\ 
    \quad set batch size = 8 in pre-training stage & 64.7 \\
    \quad set batch size = 16 in pre-training stage & 66.2 \\
    \quad set batch size = 32 in pre-training stage & 67.1 \\
    \quad set batch size = 128 in pre-training stage & 67.6 \\ \hline
    \textbf{\textit{Vocabulary compression}} & $T_1$-$S_1$ \\ 
    GLMD & \bf 67.4 \\ 
    \quad set $r_v=0.75$ & 67.2 \\ 
    \quad set $r_v=0.25$ & 65.8 \\ 
    \quad set $r_v=0.1$ & 64.6 \\
    \hline
\end{tabular}
}
\caption{Hyper-parameter analysis. All combinations use the best hyperparameters of GLMD$_{\mathrm{-vc}}$.}
\label{tab:analysis-hyper}
\end{table}

\subsection{Hyper-parameter Analysis}
In Table~\ref{tab:analysis-hyper}, we analyze the impact of these hyperparameters on performance: $\uptau$ in $\mathcal{L}^{\prime}_{\mathrm{sp}}$ and $\mathcal{L}_{\mathrm{st}}$, batch size in pre-training stage, and $r_v$ in GLMD$_{\mathrm{-vc}}$. We find that the temperature hyperparameter ($\uptau$) has a significant impact on the performance of $\mathcal{L}^{\prime}_{\mathrm{sp}}$ (rows 4-7) but little effect on $\mathcal{L}_{\mathrm{st}}$ (rows 8-14). 
Similarly, in $\mathcal{L}^{\prime}_{\mathrm{sp}}$, we observe that the batch size during the pre-training stage is roughly proportional to performance (rows 15-18). The compression ratio ($r_v$) in our vocabulary compression strategy (rows 21-23) also follows this trend as a higher $r_v$ results in more parameters being retained. It is worth noting that the teacher models ($T_1$ and $T_2$) required a batch size of 1024 during the pre-training process, which is significantly larger than the batch size we used in distillation. 

\subsection{Limitation}
Due to limitations in time and computational resources, we limited our experiments to using GLM and SuperGLUE benchmark\footnote{Given the requirement for grid search and seed averaging, we have run over a thousand SuperGLUE averages.}. While transformer-based language models and the SuperGLUE benchmark are representative, further validation is necessary when applied to a wider range of models and tasks. 
Additionally, we found that the performance of GLMD$_{\mathrm{-vc}}$ (10B→2B) at 85.28\% was marginally lower than that of GLM-2B at 85.91\%. However, it's noteworthy that GLM-2B leverages a substantially greater scale in the pre-training stage with a batch size, iterations, and GPU count of 7168, 17k, and 224 respectively, far exceeding the respective parameters of 64, 15k, and 8 employed by GLMD$_{\mathrm{-vc}}$ (10B→2B) in its distillation during the pre-training stage.
We plan to further investigate these potential limitations in future work.

\section{Conclusions}
In this paper, we introduce a general language model distillation method called GLMD. GLMD has two main advantages: improving distillation performance without relying on intermediate layer features and hard labels and reducing vocabulary parameters without reducing dimensions. We also had two important findings: distillation of intermediate layer features is unnecessary, and a vocabulary compression strategy that reduces the number of tokens is feasible and can be combined with a method that reduces dimensions. In the future, we plan to explore model distillation on a 100B-scale and apply it to more real-world scenarios.

\section*{Ethical Statement}
This paper aims to compress language models using knowledge distillation methods, and the proposed method do not raise ethical problems or potential biases. All language models, baselines, and datasets used in this work are publicly available and widely used.

\section*{Acknowledgements}
This work is supported by Technology and Innovation Major Project of the Ministry of Science and Technology of China under Grant 2020AAA0108400 and 2020AAA0108402, the Natural Science Foundation of China under Grant No. 61836013, the Major Program of the National Social Science Foundation of China under Grant No. 18ZDA032, and funds from CCF-Zhipu.AI and Beijing Academy of Artificial Intelligence (BAAI). The GPUs used are sponsored by Zhipu.AI.

\bibliography{acl2023}
\bibliographystyle{acl_natbib}

\appendix

\begin{table*}
\centering
\resizebox{\linewidth}{!}{
\begin{tabular}{l|ccccc|ccccc|ccc} \hline
    \multirow{2}{*}{\textbf{Methods}} & \multicolumn{5}{|c|}{\textbf{Pre-training}} & \multicolumn{5}{|c|}{\textbf{Task-specific}} & \multicolumn{3}{c}{\textbf{Task-specific 2}} \\ \cline{2-14}
     & \textbf{Emb} & \textbf{Att} & \textbf{HS} & \textbf{Soft} & \textbf{Hard} & \textbf{Emb} & \textbf{Att} & \textbf{HS} & \textbf{Soft} & \textbf{Hard} & \textbf{HS} & \textbf{Soft} & \textbf{Hard} \\ \hline
    KD \citep{RN4609} & \multicolumn{5}{|c|}{random parameters} &  &  &  & $\checkmark$ & $\checkmark$ &  &  &  \\ 
    PD \citep{RN6581} &  &  &  &  & $\checkmark$ &  &  &  & $\checkmark$ &  &  &  &  \\ 
    PKD \citep{RN7494} & \multicolumn{5}{|c|}{truncated teacher parameters} &  &  & $\checkmark$ & $\checkmark$ & $\checkmark$ &  &  &  \\ 
    DistilBERT \citep{RN6418} &  &  & $\checkmark$ & $\checkmark$ & $\checkmark$ &  &  &  &  & $\checkmark$ &  &  &  \\ 
    Theseus \citep{RN7113} & \multicolumn{5}{|c|}{truncated teacher parameters} &  &  & $\checkmark$ &  & $\checkmark$ &  &  &  \\ 
    TinyBERT \citep{RN7482} & $\checkmark$ & $\checkmark$ & $\checkmark$ &  &  & $\checkmark$ & $\checkmark$ & $\checkmark$ &  &  &  & $\checkmark$ &  \\
    SID \citep{RN7487} &  &  &  &  & $\checkmark$ &  & $\checkmark$ & $\checkmark$ & $\checkmark$ &  &  &  &  \\ 
    MobileBERT \citep{RN7111} & $\checkmark$ & $\checkmark$ & $\checkmark$ &  & $\checkmark$ &  &  &  &  & $\checkmark$ &  &  &  \\ 
    MiniLM \citep{RN7471} &  & $\checkmark$ &  &  &  &  &  &  &  & $\checkmark$ &  &  &  \\ 
    MiniLMv2 \citep{RN7385} &  & $\checkmark$ &  &  &  &  &  &  &  & $\checkmark$ &  &  &  \\ 
    ALP-KD \citep{RN7406} & \multicolumn{5}{|c|}{truncated teacher parameters} &  &  & $\checkmark$ & $\checkmark$ & $\checkmark$ &  &  &  \\ 
    LRC-BERT \citep{RN7437} &  &  &  &  & $\checkmark$ &  &  & $\checkmark$ &  &  & $\checkmark$ & $\checkmark$ & $\checkmark$ \\ 
    Annealing-KD \citep{RN5665} &  &  &  &  & $\checkmark$ &  &  &  & $\checkmark$ &  &  &  & $\checkmark$ \\ 
    CKD \citep{RN7407} &  &  &  &  & $\checkmark$ & $\checkmark$ &  & $\checkmark$ & $\checkmark$ & $\checkmark$ &  &  &  \\ 
    Universal-KD \citep{RN7376} & \multicolumn{5}{|c|}{truncated teacher parameters} &  &  & $\checkmark$ & $\checkmark$ &  &  &  & $\checkmark$ \\ 
    DIITO \citep{RN7325} &  &  & $\checkmark$ & $\checkmark$ & $\checkmark$ &  &  &  &  & $\checkmark$ &  &  &  \\ 
    Continuation-KD \citep{Continuation} &  &  &  &  & $\checkmark$ &  &  &  & $\checkmark$ & $\checkmark$ &  &  &  \\ 
    RAIL-KD \citep{RN7350} & \multicolumn{5}{|c|}{same as DistilBERT} &  &  & $\checkmark$ & $\checkmark$ & $\checkmark$ &  &  &  \\ 
    MGSKD \citep{RN7341} & \multicolumn{5}{|c|}{same as TinyBERT} & $\checkmark$ &  & $\checkmark$ &  &  &  & $\checkmark$ &  \\ 
    TMKD \citep{RN6109} &  &  &  & $\checkmark$ &  &  &  &  & $\checkmark$ & $\checkmark$ &  &  &  \\
    MT-BERT \citep{RN7379} & \multicolumn{5}{|c|}{truncated teacher parameters} &  &  & $\checkmark$ & $\checkmark$ & $\checkmark$ &  &  &  \\ 
    RL-KD \citep{RN7370} & \multicolumn{5}{|c|}{truncated teacher parameters} &  &  &  & $\checkmark$ & $\checkmark$ &  & $\checkmark$ & $\checkmark$ \\ 
    Uncertainty \citep{RN7419} & \multicolumn{5}{|c|}{truncated teacher parameters} &  &  &  & $\checkmark$ & $\checkmark$ &  &  &  \\
    TAKD \citep{RN7475} & \multicolumn{5}{|c|}{truncated teacher parameters} &  &  &  & $\checkmark$ & $\checkmark$ &  &  &  \\ 
    DGKD \citep{RN7395} & \multicolumn{5}{|c|}{truncated teacher parameters} &  &  &  & $\checkmark$ & $\checkmark$ &  &  &  \\ 
    \hline
\end{tabular}
}
\caption{The features used in the distillation process for the baselines implemented in GLM. A model undergoes up to three training processes (pre-training, task-specific, task-specific 2). Emb, Att, HS, Soft, and Hard denote the output of the embedding layer, attention layer (including query, key, and value), hidden state, soft labels, and hard labels, respectively.}
\label{tab:appendix:baselines}
\end{table*}

\section{Implementation Details}
\label{sec:appendix:implementation}
In this section, we provide a detailed overview of all baselines and hyperparameters for the benefit of researchers interested in a deeper analysis.

\subsection{Baselines}
\label{sec:appendix:baselines}
\begin{figure}[h]
  \centering
  \includegraphics[width=\linewidth,scale=1]{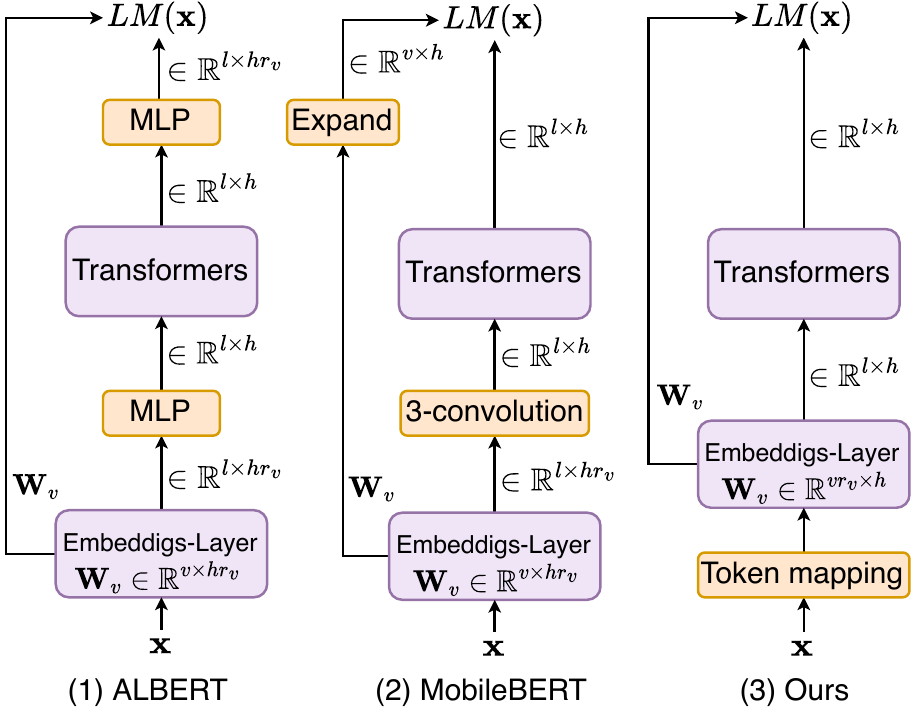}
  \caption{Details of three types of vocabulary compression strategies on the student. 
  Where $l$ denotes the length of the input sequence $\mathbf x$ (token ids) , and $r_v$ denotes the compression rate of the vocabulary. 
  It can be seen that our vocabulary compression method is characterized by reducing the number $v$ of tokens rather than dimensions $h$. 
  \label{fig:vocab}}
\end{figure}

In Table~\ref{tab:appendix:baselines}, we show the differences between 25 baseline methods in the features used. Using only hard labels for the training process is equivalent to pre-training or fine-tuning without distillation. Of these methods, 22 are specifically designed for language models, while the remaining 3 (KD, TAKD, and DGKD) are from computer vision. Figure~\ref{fig:vocab} illustrates the differences between our vocabulary compression strategy and the other two strategies. Next, we provide a brief overview of these methods, as well as some strategies we adopted and adaptations for GLM.

\textbf{KD} \citep{RN4609} was originally from computer vision and was not designed for the pre-training stage. We used randomly initialized parameters during the pre-training stage.

\textbf{PD} \citep{RN6581} removed the use of hard labels from KD, but used a pre-trained student model to initialize the student for the task-specific stage. The same hyperparameters are used for pre-training of the student model, regardless of whether distillation is performed.

\textbf{PKD} \citep{RN7494} was based on KD and added distillation loss for the [CLS] token in the intermediate layers. It was the first approach to initialize the student model for the task-specific stage by assigning some of the fine-tuned teacher's parameters to the student.

\textbf{DistilBERT} \citep{RN6418} was the first approach to use distillation during the pre-training stage and only require fine-tuning during the task-specific stage.

\textbf{Theseus} \citep{RN7113} implemented distillation by continuously replacing intermediate layers in the teacher with smaller intermediate layers.

\textbf{TinyBERT} \citep{RN7482} was the first approach to use distillation during both the pre-training and task-specific stages. We did not use data augmentation here.

\textbf{SID} \citep{RN7487} gradually increased the the number of layers for distillation as the number of epochs increased. We used the Exp3.4 strategy from the original paper.

\textbf{MobileBERT} \citep{RN7111} implemented a large-scale reduction of model parameters without reducing the number of model layers using an inverted-bottleneck structure. Since it required modifying the teacher's structure, we spent a week using the same hyperparameters as GLM-Large to pre-train an inverted-bottleneck structure of GLM-Large on 16 NVIDIA A100 GPUs. We used the PKT (progressive knowledge transfer) strategy from the original paper.

\textbf{MiniLM} \citep{RN7471} distilled the attention probability matrix and the value matrix of the final layer transformer during the pre-training stage.

\textbf{MiniLMv2} \citep{RN7385} replaced the attention probability matrix from MiniLM with query and key matrices, and modified the distillation of the final layer to other layers.

\textbf{ALP-KD} \citep{RN7406} fused the features of all layers in the teacher model through an attention mechanism, allowing each layer in the student model to capture information from all layers in the teacher.

\textbf{LRC-BERT} \citep{RN7437} constructed a loss on intermediate layer features based on contrastive learning, causing the intermediate layer features of the teacher model for other samples in the same batch to be dissimilar to the intermediate layer features of the student model for the current sample. We did not use gradient perturbation as in the original work.

\textbf{Annealing-KD} \citep{RN5665} gradually increased the weight of teacher features during the process of distilling soft targets.

\textbf{Continuation-KD} \citep{Continuation} built upon Annealing-KD by merging two training processes at the task-specific stage, resulting in the weight of hard targets increasing with the number of iterations. In addition, soft targets were not used when the value of the soft target loss was relatively small.

\textbf{CKD} \citep{RN7407} used the distance between any two or three tokens in the hidden state features as context features for the teacher and student, and then the distance between the teacher and student on these context features was used as the loss. CKD proposed task-agnostic and task-specific distillation losses, and we used task-specific distillation loss.

\textbf{Universal-KD} \citep{RN7376} used a similar attention mechanism to ALP-KD, but applied an additional linear transformation to the intermediate layer features to ensure consistency in the hidden state dimensions of the teacher and student. The original paper provided three strategies for constructing the loss, and we adopted Universal-KD$^{(IL)}$.

\textbf{DIITO} \citep{RN7325} allowed the student model to learn counterfactual outputs by exchanging intermediate layer features between different samples. This process required two forward propagations per batch, the first to extract intermediate layer features and the second to exchange them. The original paper provided multiple strategies for aligning and exchanging the intermediate layer features, and we adopted $\text{DIITO}_{\text {FULL}}+\mathcal{L}_{\text{Cos}}^{\text{DIITO}}$.

\textbf{RAIL-KD} \citep{RN7350} randomly used different intermediate layers of the teacher for distillation in each epoch of training in order to improve the generalization ability of the student model. It used a pre-trained distilled model of DistilBERT to initialize the task-specific stage. In cases where initialization with DistilBERT was not possible due to dimensional constraints (e.g. $T_1$-$S_1$ and $T_2$-$S_2$), we used MiniLM for initialization.

\textbf{MGSKD} \citep{RN7341}, based on CKD, used avg pooling to transform the hidden state features into features with three levels of granularity (token, span, sample) and constructed the loss on different layers using these granularities separately. For span representation, we randomly selected the token spans whose start positions and lengths are sampled from some distributions.

\textbf{TMKD} \citep{RN6109} introduced the multi-teacher method for language model distillation for the first time, with the aim of making the output of the student model as close as possible to the output of all the teacher models. There are two differences in our implementation compared to the original method: (1) We were unable to implement the multi-header layer, which transforms the output of the student model, due to the differences between GLM and BERT. (2) Since the original pre-training data is not publicly available, we used the same pre-trained corpus as other methods.

\textbf{MT-BERT} \citep{RN7379} first used co-finetuning to fine-tune all the teachers simultaneously, and then used the reciprocal of each teacher's loss on the task as the weight for the loss between each teacher and the student. Due to the differences between GLM and BERT, the use of co-finetuning significantly degraded the performance of the teacher models, so we did not use co-finetuning.

\textbf{RL-KD} \citep{RN7370} used reinforcement learning to select appropriate teachers for distillation at each iteration, and the final loss was the average of the loss between each selected teacher and the student. We used the $reward_1$ from the original paper as the method for calculating the reward.

\textbf{Uncertainty} \citep{RN7419} used the entropy of the student's predicted results as a criterion for selecting the teacher at each iteration. The lower the entropy, the more confident the student was and the more it learned from the larger scale teacher, a process referred to as dynamic teacher adoption. We employed the hard selection strategy from the original paper.

\textbf{TAKD} \citep{RN7475}, for the first time, used a teacher assistant approach in which the teacher was distilled to a mid-sized teacher assistant before being distilled to the student, rather than distilling the teacher directly to the student.

\textbf{DGKD} \citep{RN7395}, building upon TAKD, used all previously distilled teachers and assistants to distill the current model. It randomly discarded teachers or assistants at each iteration to serve as a regularizer.

\subsection{Hyperparameters}
\label{sec:appendix:hyper}
To ensure the reproducibility of all methods, we present in Table~\ref{tab:appendix:hyper} the learning rates and batch sizes for each method on each dataset in the SuperGLUE benchmark, including the hyperparameters obtained via grid search. Table~\ref{tab:appendix:other-hyper} further shows the additional hyperparameters for the task-specific stage, which follow the settings of GLM.

\begin{table*}
\centering
\resizebox{\linewidth}{!}{
\begin{tabular}{lc|cccccccc} \hline
    \multirow{2}{*}{\textbf{Methods}} & \multirow{2}{*}{\textbf{Size}} & \textbf{ReCoRD} & \textbf{COPA} & \textbf{WSC} & \textbf{RTE} & \textbf{BoolQ} & \textbf{WiC} & \textbf{CB} & \textbf{MultiRC} \\ \cline{3-10}
     &  & \textbf{bs/lr}& \textbf{bs/lr}& \textbf{bs/lr}& \textbf{bs/lr}& \textbf{bs/lr}& \textbf{bs/lr}& \textbf{bs/lr}& \textbf{bs/lr} \\ \hline
    \multirow{3}{*}{GLM (teacher)} & $T_1$ & \multicolumn{8}{|c}{\multirow{2}{*}{bs (batch size) = 16, lr (learning rate) = 1E-5}} \\
     & $T_2$ &  &  &  &  &  &  &  &  \\
     & $T_3$ & \multicolumn{8}{|c}{bs = 64, lr = 1E-5} \\
    PKD & \multirow{6}{*}{$T_1$-$S_2$} & 32/2E-05 & 32/2E-05 & 16/2E-05 & 32/5E-06 & 16/1E-05 & 16/5E-06 & 16/2E-05 & 32/2E-05 \\ 
    DistilBERT &  & 16/1E-05 & 16/2E-05 & 16/1E-05 & 16/5E-06 & 32/2E-05 & 32/2E-05 & 32/2E-05 & 16/1E-05 \\ 
    Theseus &  & 32/2E-05 & 16/1E-05 & 16/1E-05 & 32/1E-05 & 16/1E-05 & 32/1E-05 & 16/2E-05 & 32/5E-06 \\ 
    SID &  & 16/2E-05 & 32/5E-06 & 16/5E-06 & 16/2E-05 & 16/2E-05 & 16/2E-05 & 16/1E-05 & 16/2E-05 \\ 
    ALP-KD &  & 16/2E-05 & 16/1E-05 & 16/2E-05 & 16/2E-05 & 16/2E-05 & 32/2E-05 & 16/2E-05 & 32/2E-05 \\ 
    DIITO &  & 16/5E-06 & 32/1E-05 & 16/2E-05 & 16/1E-05 & 16/2E-05 & 16/1E-05 & 16/1E-05 & 16/5E-06 \\ 
    MobileBERT & $T_1$-$S_1$ & 16/1E-05 & 16/1E-05 & 32/2E-05 & 32/2E-05 & 32/2E-05 & 32/1E-05 & 32/2E-05 & 16/5E-06 \\ \hline
    \multirow{3}{*}{KD} & $T_1$-$S_1$ & 16/2E-05 & 16/2E-05 & 16/2E-05 & 32/5E-06 & 32/1E-05 & 32/5E-06 & 16/2E-05 & 16/5E-06 \\ 
     & $T_1$-$S_2$ & 16/5E-06 & 16/2E-05 & 16/1E-05 & 16/2E-05 & 16/2E-05 & 16/5E-06 & 16/2E-05 & 16/5E-06 \\ 
     & $T_2$-$S_2$ & 16/2E-05 & 16/1E-05 & 16/1E-05 & 32/1E-05 & 16/1E-05 & 16/5E-06 & 16/1E-05 & 32/5E-06 \\  \hline
    \multirow{3}{*}{PD} & $T_1$-$S_1$ & 32/2E-05 & 16/1E-05 & 16/2E-05 & 16/1E-05 & 16/2E-05 & 16/5E-06 & 16/2E-05 & 32/1E-05  \\ 
     & $T_1$-$S_2$ & 16/1E-05 & 32/5E-06 & 16/2E-05 & 16/1E-05 & 32/1E-05 & 16/5E-06 & 16/2E-05 & 16/5E-06 \\ 
     & $T_2$-$S_2$ & 16/1E-05 & 32/5E-06 & 16/2E-05 & 16/1E-05 & 16/2E-05 & 16/5E-06 & 32/2E-05 & 16/1E-05 \\  \hline
    \multirow{4}{*}{TinyBERT} & $T_1$-$S_1$ & 32/1E-05 & 16/5E-06 & 16/2E-05 & 32/1E-05 & 32/2E-05 & 16/2E-05 & 16/1E-05 & 32/2E-05 \\ 
     & $T_1$-$S_2$ & 32/1E-05 & 16/5E-06 & 32/5E-06 & 16/2E-05 & 16/1E-05 & 16/5E-06 & 16/1E-05 & 16/1E-05 \\ 
     & $T_2$-$S_2$ & 32/1E-05 & 32/5E-06 & 32/2E-05 & 32/1E-05 & 16/1E-05 & 16/1E-05 & 16/2E-05 & 16/2E-05 \\
     & $T_3$-$S_3$ & \multicolumn{8}{|c}{same as GLM (teacher, $T_1$)} \\  \hline
    \multirow{3}{*}{MiniLM} & $T_1$-$S_1$ & 16/2E-05 & 32/2E-05 & 16/1E-05 & 16/5E-06 & 32/2E-05 & 32/5E-06 & 16/2E-05 & 16/1E-05  \\ 
     & $T_1$-$S_2$ & 16/2E-05 & 32/1E-05 & 32/2E-05 & 32/1E-05 & 16/1E-05 & 16/1E-05 & 32/1E-05 & 32/2E-05 \\ 
     & $T_2$-$S_2$ & 16/1E-05 & 32/5E-06 & 16/2E-05 & 32/1E-05 & 16/5E-06 & 16/5E-06 & 16/1E-05 & 16/5E-06 \\  \hline
    \multirow{3}{*}{MiniLMv2} & $T_1$-$S_1$ & 16/1E-05 & 16/1E-05 & 32/1E-05 & 32/1E-05 & 32/2E-05 & 32/1E-05 & 16/2E-05 & 16/5E-06  \\ 
     & $T_1$-$S_2$ & 16/1E-05 & 16/1E-05 & 16/5E-06 & 32/2E-05 & 16/2E-05 & 32/2E-05 & 16/1E-05 & 16/1E-05 \\ 
     & $T_2$-$S_2$ & 16/1E-05 & 16/2E-05 & 16/2E-05 & 32/5E-06 & 16/2E-05 & 16/2E-05 & 16/2E-05 & 32/2E-05 \\  \hline
    \multirow{3}{*}{LRC-BERT} & $T_1$-$S_1$ & 16/2E-05 & 16/2E-05 & 32/2E-05 & 16/5E-06 & 16/2E-05 & 16/5E-06 & 16/5E-06 & 16/2E-05  \\ 
     & $T_1$-$S_2$ & 16/2E-05 & 32/1E-05 & 16/2E-05 & 32/1E-05 & 16/2E-05 & 16/5E-06 & 16/2E-05 & 16/5E-06 \\ 
     & $T_2$-$S_2$ & 16/2E-05 & 32/5E-06 & 32/5E-06 & 32/5E-06 & 16/2E-05 & 32/5E-06 & 32/2E-05 & 16/2E-05 \\  \hline
    \multirow{3}{*}{Annealing-KD} & $T_1$-$S_1$ & 16/2E-05 & 32/5E-06 & 32/2E-05 & 32/2E-05 & 16/1E-05 & 32/5E-06 & 32/2E-05 & 32/5E-06 \\ 
     & $T_1$-$S_2$ & 16/2E-05 & 16/5E-06 & 16/2E-05 & 16/2E-05 & 16/2E-05 & 32/5E-06 & 16/1E-05 & 32/5E-06 \\ 
     & $T_2$-$S_2$ & 16/2E-05 & 16/1E-05 & 32/2E-05 & 32/5E-06 & 32/2E-05 & 32/2E-05 & 32/1E-05 & 16/5E-06 \\  \hline
    \multirow{3}{*}{CKD} & $T_1$-$S_1$ & 32/1E-05 & 16/5E-06 & 16/2E-05 & 16/5E-06 & 16/2E-05 & 32/5E-06 & 32/2E-05 & 16/5E-06  \\ 
     & $T_1$-$S_2$ & 32/2E-05 & 16/2E-05 & 16/5E-06 & 16/1E-05 & 16/2E-05 & 16/1E-05 & 16/1E-05 & 32/2E-05 \\ 
     & $T_2$-$S_2$ & 16/1E-05 & 32/1E-05 & 16/1E-05 & 32/1E-05 & 16/2E-05 & 32/1E-05 & 16/2E-05 & 32/2E-05 \\  \hline
    \multirow{3}{*}{Universal-KD} & $T_1$-$S_1$ & 16/2E-05 & 32/2E-05 & 32/2E-05 & 32/5E-06 & 16/2E-05 & 16/1E-05 & 32/2E-05 & 16/2E-05  \\ 
     & $T_1$-$S_2$ & 32/2E-05 & 32/5E-06 & 32/5E-06 & 32/1E-05 & 32/5E-06 & 16/5E-06 & 16/1E-05 & 16/1E-05 \\ 
     & $T_2$-$S_2$ & 16/5E-06 & 32/1E-05 & 32/2E-05 & 16/2E-05 & 16/1E-05 & 16/5E-06 & 32/1E-05 & 32/2E-05 \\  \hline
    \multirow{3}{*}{Continuation-KD} & $T_1$-$S_1$ & 16/2E-05 & 32/5E-06 & 16/2E-05 & 32/2E-05 & 32/2E-05 & 32/2E-05 & 16/2E-05 & 32/1E-05 \\ 
     & $T_1$-$S_2$ & 16/2E-05 & 32/1E-05 & 16/1E-05 & 16/1E-05 & 16/2E-05 & 32/1E-05 & 16/1E-05 & 16/5E-06 \\ 
     & $T_2$-$S_2$ & 16/1E-05 & 16/1E-05 & 16/2E-05 & 16/2E-05 & 16/1E-05 & 16/5E-06 & 16/1E-05 & 32/5E-06 \\  \hline
    \multirow{3}{*}{RAIL-KD} & $T_1$-$S_1$ & 16/1E-05 & 32/1E-05 & 32/2E-05 & 32/1E-05 & 16/1E-05 & 32/5E-06 & 16/2E-05 & 32/2E-05  \\ 
     & $T_1$-$S_2$ & 16/1E-05 & 16/1E-05 & 16/2E-05 & 16/5E-06 & 32/2E-05 & 16/1E-05 & 32/1E-05 & 32/2E-05 \\ 
     & $T_2$-$S_2$ & 32/2E-05 & 16/2E-05 & 32/2E-05 & 16/5E-06 & 16/1E-05 & 16/5E-06 & 16/2E-05 & 16/1E-05 \\  \hline
    \multirow{3}{*}{MGSKD} & $T_1$-$S_1$ & 32/1E-05 & 16/5E-06 & 32/2E-05 & 32/1E-05 & 16/1E-05 & 32/1E-05 & 32/5E-06 & 32/5E-06  \\ 
     & $T_1$-$S_2$ & 16/5E-06 & 16/2E-05 & 32/2E-05 & 16/5E-06 & 16/5E-06 & 16/1E-05 & 32/2E-05 & 32/5E-06 \\ 
     & $T_2$-$S_2$ & 32/2E-05 & 32/5E-06 & 32/5E-06 & 32/5E-06 & 16/2E-05 & 16/5E-06 & 16/2E-05 & 16/2E-05 \\  \hline
    \multirow{4}{*}{GLMD$_{\mathrm{-vc}}$} & $T_1$-$S_1$ & 16/2E-05 & 32/2E-05 & 32/2E-05 & 16/5E-06 & 16/2E-05 & 16/2E-05 & 32/2E-05 & 16/1E-05 \\ 
     & $T_1$-$S_2$ & 16/2E-05 & 16/5E-06 & 32/2E-05 & 16/2E-05 & 16/2E-05 & 16/2E-05 & 32/2E-05 & 16/5E-06 \\ 
     & $T_2$-$S_2$ & 16/1E-05 & 32/1E-05 & 32/2E-05 & 32/2E-05 & 16/2E-05 & 16/2E-05 & 32/2E-05 & 16/5E-06 \\
     & $T_3$-$S_3$ & \multicolumn{8}{|c}{same as GLM (teacher, $T_1$)} \\  \hline
    \multirow{3}{*}{GLMD} & $T_1$-$S_1$ & \multicolumn{8}{|c}{same as GLMD$_{\mathrm{-vc}}$ ($T_1$-$S_1$)} \\
     & $T_1$-$S_2$ & \multicolumn{8}{|c}{same as GLMD$_{\mathrm{-vc}}$ ($T_1$-$S_2$)} \\ 
     & $T_2$-$S_2$ & \multicolumn{8}{|c}{same as GLMD$_{\mathrm{-vc}}$ ($T_2$-$S_2$)} \\ \hline
    TMKD & \multirow{4}{*}{$(T_1,T_2)$-$S_2$} & \multicolumn{8}{|c}{\multirow{4}{*}{same as GLM (teacher, $T_1$)}} \\
    MT-BERT &  &  &  &  &  &  &  &  &  \\ 
    RL-KD &  &  &  &  &  &  &  &  &  \\
    Uncertainty &  &  &  &  &  &  &  &  &  \\ \hline
    TAKD & \multirow{2}{*}{$T_2$-$A_1$-$A_2$-$S_2$} & \multicolumn{8}{|c}{\multirow{2}{*}{same as KD ($T_1$-$S_2$)}} \\
    DGKD &  &  &  &  &  &  &  &  &  \\ \hline
    GLMD$_{\mathrm{-vc+mo}}$ & \multirow{3}{*}{$T_1$-$S_1$} & \multicolumn{8}{|c}{\multirow{3}{*}{same as GLMD ($T_1$-$S_1$)}} \\
    GLMD$_{\mathrm{-vc+al}}$ &  &  &  &  &  &  &  &  &  \\
    GLMD$_{\mathrm{+al}}$ &  &  &  &  &  &  &  &  &  \\
    \hline
\end{tabular}
}
\caption{Hyperparameters for all methods in Table~\ref{tab:results} on the 8 datasets of the SuperGLUE benchmark.}
\label{tab:appendix:hyper}
\end{table*}

\begin{table*}
\centering
\resizebox{0.75\linewidth}{!}{
\begin{tabular}{l|cccccccc} \hline
	\textbf{Hyperparameters} & \textbf{ReCoRD} & \textbf{COPA} & \textbf{WSC} & \textbf{RTE} & \textbf{BoolQ} & \textbf{WiC} & \textbf{CB} & \textbf{MultiRC} \\ \hline
    Sequence length & 512 & 256 & 128 & 256 & 256 & 256 & 256 & 512 \\ 
    Epochs & 5 & 50 & 50 & 50 & 20 & 30 & 50 & 15 \\ 
    Dropout & \multicolumn{8}{|c}{0.1} \\ 
    Attention Dropout & \multicolumn{8}{|c}{0.1} \\ 
    Warmup Ration & \multicolumn{8}{|c}{0.1} \\ 
    Weight Decay & \multicolumn{8}{|c}{0.1} \\ 
    Learning Rate Decay & \multicolumn{8}{|c}{Linear} \\ 
    Adam $\epsilon$ & \multicolumn{8}{|c}{1E-8} \\ 
    Adam $\beta_1$ & \multicolumn{8}{|c}{0.9} \\ 
    Adam $\beta_2$ & \multicolumn{8}{|c}{0.999} \\ 
    Gradient Clipping & \multicolumn{8}{|c}{0.1} \\ 
    \hline
\end{tabular}
}
\caption{Other hyperparameters for the task-specific stage.}
\label{tab:appendix:other-hyper}
\end{table*}

\begin{figure*}
  \centering
  \includegraphics[width=\linewidth]{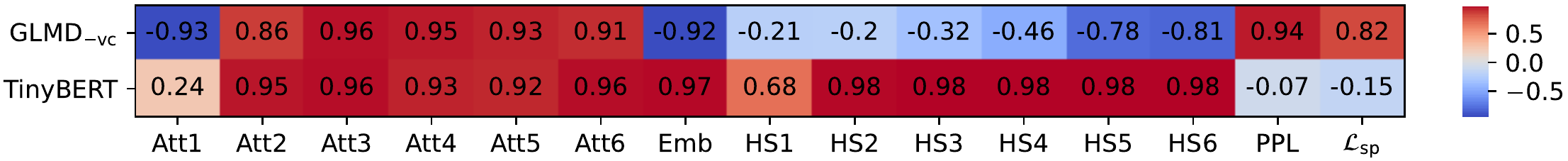}
  \caption{Spearman correlation coefficient of the loss function values with the distance between teacher and student features during pre-training stage of GLMD$_{\mathrm{-vc}}$ and TinyBERT ($T_1$-$S_2$). Att1 and HS1 denote the attention scores and hidden states, respectively, of the first layer transformer. Emb denotes the output of the embedding layer. The KL divergence with a temperature $\uptau$ of 1 is used to compute the distance between the teacher and student models in the Att, while the mean square error is used for the HS and Emb. PPL denotes the language modeling perplexity of the student on the validation set. \label{fig:appendix:glmd-tinybert-analysis}}
\end{figure*}

\section{Additional Analysis}
\label{sec:appendix:analysis}
In this section, we further explore the various factors that influence the performance of the baselines and examine the necessity of intermediate layer features.

\subsection{What Factors Affect Performance?}
In the implementation of baselines, we discover that certain methods for initializing the student parameters led to a decrease in performance, surpassing even the advantages brought about by the method innovation. Specifically, these include the following three ways: \textbf{(1) using truncated teacher parameters in the case of different dimensions of teacher and student hidden layers.} Many methods that do not distill in the pre-training stage will use the parameters of the first few layers of the fine-tuned teacher as student parameters in a task-specific stage. In the case of different dimensions of teacher and student hidden layers, we can only truncate some parameters per layer. As shown in Table~\ref{tab:results}, Universal-KD \citep{RN7376} in $T_1$-$S_2$ performs much better than $T_1$-$S_2$ and $T_2$-$S_2$, TAKD \citep{RN7475} and DGKD \citep{RN7395} also performs badly due to this reason. \textbf{(2) Using less data to pre-train a student model as initialization.} To ensure fairness, regardless of whether distillation is used, we set the batch size of all methods in the pre-training stage to 64, which is equivalent to only using one-sixteenth of the full data (batch size=1024). Some methods that use a pre-trained student as initialization for a task-specific stage may be affected by this, for example, PD \citep{RN6581}, SID \citep{RN7487}, LRC-BERT \citep{RN7437}, Annealing-KD \citep{RN5665}, Continuation-KD \citep{Continuation}, and CKD \citep{RN7407}. \textbf{(3) Randomly initializing the parameters of the student model.} As can be seen from Table~\ref{tab:results}, KD \citep{RN4609} using random initialization are obviously inferior to PD \citep{RN6581} using pre-trained students and soft labels.

The above analysis demonstrates that methods without pre-training distillation are sensitive to the initialization of the student's parameters. To achieve optimal performance, methods based on truncating teacher parameters require the hidden dimensions of the teacher and student to be identical or else, other methods would require a significant cost in pre-training a student model. Therefore, utilizing a subset of the corpus for knowledge distillation during the pre-training stage is a more favorable option.

\subsection{Why are Intermediate Layers not Necessary?}
In Section~\ref{analysis:why}, we have verified that the $\mathcal{L}^{\prime}_{\mathrm{sp}}$ in GLMD$_{\mathrm{-vc}}$ can enable the model to spontaneously learn intermediate layer features that should be similar to those of the teacher. We further validate in Figure~\ref{fig:appendix:glmd-tinybert-analysis} that training with a loss function focused on the intermediate layer features does not lead to a reduction in $\mathcal{L}^{\prime}_{\mathrm{sp}}$ and cannot even lower the perplexity (PPL) of language modeling. In the process of distillation using GLMD$_{\mathrm{-vc}}$ and TinyBERT methods, we have quantified the spearman correlation between the distance of student and teacher features, including the perplexity of the student model on the validation set, and the loss function values. We observe that there is no correlation between the loss function values of TinyBERT and $\mathcal{L}^{\prime}_{\mathrm{sp}}$, nor with the perplexity of the validation set. This suggests that we may not require a significant inductive bias towards the intermediate layer features.

\begin{table*}
\centering
\resizebox{\linewidth}{!}{
\begin{tabular}{lc|ccccccccc} \hline
    \multirow{2}{*}{\textbf{Methods}} & \multirow{2}{*}{\textbf{Size}} & \textbf{ReCoRD} & \textbf{COPA} & \textbf{WSC} & \textbf{RTE} & \textbf{BoolQ} & \textbf{WiC} & \textbf{CB} & \textbf{MultiRC} & \multirow{2}{*}{\textbf{avg}} \\ \cline{3-10}
     &  & \textbf{F1/Acc.}& \textbf{Acc.}& \textbf{Acc.}& \textbf{Acc.}& \textbf{Acc.}& \textbf{Acc.}& \textbf{F1/Acc.}& \textbf{F1$_a$/EM} & \\ \hline
    \multirow{3}{*}{GLM (teacher)} & $T_1$ & 72.80/72.17 & 66.00 & 77.88 & 72.92 & 79.39 & 66.14 & 88.19/91.07 & 72.32/26.34 & 71.72 \\
     & $T_2$ & 80.08/79.54 & 78.00 & 81.73 & 79.78 & 82.63 & 70.06 & 86.33/89.29 & 76.39/37.67 & 77.11 \\
     & $T_3$ & 94.56/94.13 & 97.00 & 95.19 & 90.61 & 88.81 & 74.61 & 96.05/94.64 & 88.10/63.38 & 88.96 \\
    PKD & \multirow{6}{*}{$T_1$-$S_2$} & 61.77/60.99 & 60.00 & 65.38 & 68.83 & 77.73 & 65.78 & 82.76/85.12 & 69.99/22.67 & 66.17 \\ 
    DistilBERT &  & 59.79/59.05 & 65.00 & 68.59 & 60.89 & 73.39 & 60.34 & 77.48/83.33 & 66.98/17.38 & 63.78 \\ 
    Theseus &  & 57.07/56.33 & 61.67 & 66.35 & 68.11 & 77.81 & 64.37 & 89.14/87.50 & 69.08/21.79 & 66.09 \\ 
    SID &  & 27.17/26.19 & 65.00 & 65.06 & 58.12 & 69.33 & 57.16 & 51.02/73.81 & 59.26/14.55 & 55.08 \\ 
    ALP-KD &  & 57.72/56.90 & 60.67 & 64.74 & 68.11 & 77.20 & 64.79 & 74.82/79.76 & 68.21/19.90 & 64.27 \\ 
    DIITO &  & 63.71/63.00 & 72.00 & 69.23 & 65.46 & 75.46 & 60.76 & 86.75/85.12 & 66.28/17.63 & 66.77 \\ 
    MobileBERT & $T_1$-$S_1$ & 59.29/58.61 & 65.33 & 68.59 & 58.97 & 74.61 & 63.85 & 86.65/88.69 & 66.87/19.41 & 65.14 \\ \hline
    \multirow{3}{*}{KD} & $T_1$-$S_1$ & 22.43/21.74 & 58.67 & 64.74 & 54.75 & 65.98 & 56.69 & 65.56/72.02 & 46.20/2.66 & 52.02 \\ 
     & $T_1$-$S_2$ & 22.66/21.99 & 61.67 & 63.46 & 54.63 & 66.07 & 57.05 & 61.75/72.02 & 51.98/2.41 & 52.41 \\ 
     & $T_2$-$S_2$ & 23.07/22.38 & 59.33 & 65.71 & 54.15 & 66.12 & 56.79 & 69.19/74.40 & 56.56/1.50 & 53.21 \\  \hline
    \multirow{3}{*}{PD} & $T_1$-$S_1$ & 46.54/45.90 & 66.33 & 67.95 & 58.48 & 69.93 & 59.67 & 81.78/80.95 & 63.91/12.42 & 61.01 \\ 
     & $T_1$-$S_2$ & 54.36/53.59 & 65.67 & 66.67 & 59.45 & 69.82 & 59.20 & 80.13/81.55 & 65.97/15.29 & 62.03 \\ 
     & $T_2$-$S_2$ & 54.00/53.22 & 65.33 & 64.74 & 60.29 & 69.94 & 58.41 & 76.66/79.76 & 66.05/15.15 & 61.39 \\  \hline
    \multirow{4}{*}{TinyBERT} & $T_1$-$S_1$ & 56.13/55.53 & 69.00 & 69.87 & 70.04 & 76.93 & 65.41 & 84.86/85.12 & 70.60/22.39 & 67.32 \\ 
     & $T_1$-$S_2$ & 65.60/64.88 & 70.33 & 75.00 & 71.96 & 77.97 & 67.87 & 89.58/89.88 & 71.37/25.74 & 70.83 \\ 
     & $T_2$-$S_2$ & 66.61/65.86 & 67.00 & 63.46 & 71.12 & 79.98 & 66.46 & 90.56/88.69 & 70.42/27.35 & 69.09 \\
     & $T_3$-$S_3$ & - & 61.00 & 65.38 & 67.87 & 74.46 & 63.32 & 70.49/78.57 & 70.76/27.39 & 65.09 \\   \hline
    \multirow{3}{*}{MiniLM} & $T_1$-$S_1$ & 51.01/50.27 & 63.67 & 60.90 & 67.63 & 73.72 & 61.29 & 68.87/77.98 & 66.93/16.19 & 61.60 \\ 
     & $T_1$-$S_2$ & 60.00/59.24 & 62.00 & 63.46 & 67.63 & 75.88 & 64.99 & 67.63/79.17 & 67.36/19.66 & 63.81 \\ 
     & $T_2$-$S_2$ & 57.07/56.39 & 62.33 & 64.42 & 66.43 & 74.00 & 60.92 & 65.38/79.76 & 66.81/17.45 & 62.44 \\  \hline
    \multirow{3}{*}{MiniLMv2} & $T_1$-$S_1$ & 51.85/51.25 & 65.00 & 60.58 & 67.63 & 75.17 & 61.70 & 65.81/79.76 & 66.26/17.98 & 62.07 \\ 
     & $T_1$-$S_2$ & 60.88/60.16 & 62.00 & 62.82 & 66.67 & 76.73 & 63.69 & 66.38/76.79 & 68.68/21.65 & 63.65 \\ 
     & $T_2$-$S_2$ & 64.08/63.29 & 59.33 & 65.71 & 66.19 & 77.26 & 65.05 & 86.84/85.71 & 68.40/21.44 & 66.05 \\  \hline
    \multirow{3}{*}{LRC-BERT} & $T_1$-$S_1$ & 40.44/39.69 & 64.33 & 66.03 & 54.87 & 68.84 & 56.74 & 78.68/80.36 & 59.66/13.61 & 58.38 \\ 
     & $T_1$-$S_2$ & 55.10/54.44 & 65.67 & 66.67 & 56.56 & 74.86 & 57.63 & 80.27/81.55 & 65.75/16.16 & 62.25 \\ 
     & $T_2$-$S_2$ & 51.83/51.17 & 66.33 & 63.78 & 58.12 & 72.02 & 59.04 & 68.08/75.00 & 66.69/21.13 & 60.78 \\  \hline
    \multirow{3}{*}{Annealing-KD} & $T_1$-$S_1$ & 49.18/48.55 & 66.33 & 65.38 & 58.97 & 69.68 & 58.36 & 82.73/81.55 & 63.96/12.91 & 61.02 \\ 
     & $T_1$-$S_2$ & 56.08/55.39 & 69.33 & 66.67 & 58.97 & 70.57 & 59.82 & 85.78/85.12 & 66.26/13.92 & 63.33 \\ 
     & $T_2$-$S_2$ & 55.57/54.89 & 69.00 & 68.59 & 58.24 & 71.48 & 58.88 & 83.11/83.33 & 66.85/13.43 & 63.10 \\  \hline
    \multirow{3}{*}{CKD} & $T_1$-$S_1$ & 48.82/48.27 & 66.33 & 64.74 & 59.57 & 70.65 & 60.76 & 87.02/85.71 & 63.72/12.28 & 61.87 \\ 
     & $T_1$-$S_2$ & 56.35/55.65 & 65.00 & 66.67 & 61.25 & 71.63 & 58.83 & 88.61/84.52 & 66.11/15.22 & 63.33 \\ 
     & $T_2$-$S_2$ & 56.29/55.57 & 65.00 & 65.71 & 58.00 & 71.39 & 58.46 & 86.17/84.52 & 66.49/14.62 & 62.55 \\  \hline
    \multirow{3}{*}{Universal-KD} & $T_1$-$S_1$ & 24.08/23.27 & 61.00 & 66.03 & 55.48 & 65.93 & 56.79 & 60.38/73.21 & 53.17/2.17 & 52.92 \\ 
     & $T_1$-$S_2$ & 58.67/57.83 & 58.67 & 66.67 & 70.16 & 77.56 & 65.52 & 87.52/85.71 & 69.96/22.63 & 66.22 \\ 
     & $T_2$-$S_2$ & 24.51/23.71 & 64.00 & 67.63 & 55.84 & 66.47 & 58.52 & 66.11/75.60 & 56.39/1.22 & 54.53 \\  \hline
    \multirow{3}{*}{Continuation-KD} & $T_1$-$S_1$ & 48.63/48.01 & 66.33 & 66.03 & 58.97 & 69.12 & 58.31 & 83.72/81.55 & 63.38/13.26 & 61.00 \\ 
     & $T_1$-$S_2$ & 55.61/54.91 & 68.67 & 64.74 & 58.72 & 71.42 & 58.25 & 85.61/83.93 & 66.64/13.33 & 62.73 \\ 
     & $T_2$-$S_2$ & 55.15/54.38 & 67.00 & 65.38 & 57.64 & 70.91 & 58.20 & 78.79/80.36 & 66.80/13.96 & 61.73 \\  \hline
    \multirow{3}{*}{RAIL-KD} & $T_1$-$S_1$ & 51.49/50.83 & 62.67 & 67.31 & 64.50 & 71.93 & 60.45 & 65.91/75.60 & 65.06/16.79 & 61.21 \\ 
     & $T_1$-$S_2$ & 59.85/59.19 & 66.67 & 70.19 & 60.53 & 69.00 & 60.34 & 78.98/83.33 & 66.55/15.60 & 63.56 \\ 
     & $T_2$-$S_2$ & 50.26/49.51 & 62.00 & 65.06 & 59.33 & 72.30 & 59.46 & 73.17/78.57 & 63.57/15.01 & 60.40 \\  \hline
    \multirow{3}{*}{MGSKD} & $T_1$-$S_1$ & 34.03/33.26 & 65.33 & 61.86 & 64.98 & 70.20 & 61.23 & 70.98/77.98 & 64.38/12.63 & 58.78 \\ 
     & $T_1$-$S_2$ & 50.29/49.49 & 65.00 & 65.06 & 65.94 & 73.31 & 63.17 & 83.89/84.52 & 67.32/15.56 & 63.50 \\ 
     & $T_2$-$S_2$ & 43.68/42.88 & 63.00 & 63.78 & 57.76 & 72.46 & 60.55 & 85.64/85.71 & 55.27/10.74 & 59.94 \\  \hline
    \multirow{4}{*}{GLMD$_{\mathrm{-vc}}$} & $T_1$-$S_1$ & 59.06/58.33 & 74.00 & 72.12 & 66.19 & 76.17 & 64.26 & 86.96/86.90 & 67.97/22.11 & \bf 67.92 \\ 
     & $T_1$-$S_2$ & 68.66/67.90 & 72.00 & 75.96 & 70.16 & 78.92 & 67.08 & 91.59/89.88 & 70.29/27.11 & \bf 71.48 \\ 
     & $T_2$-$S_2$ & 70.51/69.82 & 72.67 & 73.08 & 71.60 & 79.96 & 66.93 & 93.08/91.67 & 71.76/28.86 & 72.13 \\
     & $T_3$-$S_3$ & 89.35/88.50 & 89.00 & 86.54 & 87.36 & 86.24 & 74.14 & 100.00/100.00 & 84.78/55.30 & \bf 85.28 \\ \hline
    \multirow{3}{*}{GLMD} & $T_1$-$S_1$ & 57.98/57.23 & 71.33 & 68.27 & 65.10 & 76.41 & 63.95 & 91.99/90.48 & 68.61/21.79 & 67.39 \\
     & $T_1$-$S_2$ & 66.13/65.44 & 72.33 & 75.00 & 71.12 & 78.30 & 66.25 & 89.61/90.48 & 70.53/25.78 & 70.87 \\ 
     & $T_2$-$S_2$ & 68.97/68.32 & 75.33 & 74.36 & 71.84 & 80.37 & 65.78 & 92.90/90.48 & 71.64/28.09 & \bf 72.23 \\ \hline
    TMKD & \multirow{4}{*}{$(T_1,T_2)$-$S_2$} & 65.77/65.09 & 70.33 & 63.14 & 66.91 & 75.37 & 63.38 & 70.22/79.17 & 68.76/22.77 & 65.63 \\ 
    MT-BERT &  & 46.81/46.08 & 59.00 & 63.46 & 65.46 & 66.90 & 62.33 & 78.76/80.36 & 57.53/2.06 & 59.12 \\ 
    RL-KD &  & 59.78/58.99 & 58.33 & 66.03 & 69.07 & 77.93 & 65.78 & 76.87/82.74 & 69.24/22.21 & 65.26 \\
    Uncertainty &  & 58.52/57.67 & 59.33 & 64.10 & 70.16 & 77.55 & 65.78 & 80.85/83.33 & 69.47/22.49 & 65.39 \\ \hline
    TAKD & \multirow{2}{*}{$T_2$-$A_1$-$A_2$-$S_2$} & 25.50/24.69 & 60.33 & 66.03 & 55.11 & 66.39 & 57.94 & 76.28/76.79 & 55.90/1.50 & 54.52 \\ 
    DGKD &  & 23.68/22.96 & 61.00 & 66.99 & 55.96 & 65.71 & 58.73 & 75.45/75.60 & 48.06/1.50 & 54.00 \\ \hline
    GLMD$_{\mathrm{-vc+mo}}$ & \multirow{3}{*}{$T_1$-$S_1$} & 59.56/58.85 & 67.67 & 70.51 & 68.35 & 77.41 & 64.99 & 81.48/82.74 & 68.78/22.74 & 67.00 \\ 
    GLMD$_{\mathrm{-vc+al}}$ &  & 59.79/59.13 & 69.67 & 65.38 & 71.12 & 76.95 & 64.37 & 84.81/88.10 & 68.76/21.76 & 67.33 \\
    GLMD$_{\mathrm{+al}}$ &  & 58.74/58.06 & 70.67 & 70.19 & 69.55 & 76.82 & 63.11 & 85.18/85.71 & 68.24/22.14 & \bf 67.42 \\
    \hline
\end{tabular}
}
\caption{Detailed results for all methods in Table~\ref{tab:results} on the 8 datasets of the SuperGLUE benchmark.}
\label{tab:appendix:results}
\end{table*}

\section{Detailed Results}
\label{sec:appendix:results}
Due to space constraints, we do not present results for all datasets in the SuperGLUE benchmark in the main text but only show the averages. Table~\ref{tab:appendix:results} shows the results for all methods on each dataset in the SuperGLUE benchmark, rounded to two decimal places.

\end{document}